\documentclass[runningheads]{llncs}
\usepackage{graphicx}

\usepackage{comment}
\usepackage{amsmath}
\usepackage[para]{footmisc}
\usepackage{multirow} % required for tables
\usepackage{rotating} % for landscape tables
\usepackage[normalem]{ulem}
\useunder{\uline}{\ul}{}
\usepackage[inline]{enumitem}
\usepackage{tikz}
\usepackage{mathtools}
\newcommand\encircle[1]{%
  \tikz[baseline=(X.base)] 
    \node (X) [draw, shape=circle, inner sep=0] {\strut #1};}
\usepackage{comment}

\begin{document}
%
%\title{The Importance of Being Ordered: Entity Typing with RDF2vec Embedding Variants}
\title{Entity Type Prediction Leveraging Graph Walks and Entity Descriptions}
%\title{Entity Type Prediction Leveraging RDF2vec and Entity Descriptions}
%Mehwish (suggestion): 
%Title: "Entity Type Prediction Leveraging Graph Walks and Entity Descriptions" 
%Suggestion for method name: WAND = graph Walks And eNtity Description or GRAND = Graph walks for Rdf2vec And eNtity Descriptions? :D

\titlerunning{Entity Type Prediction Leveraging Graph Walks and Entity Descriptions}
\author{
Russa Biswas\inst{1,2}\thanks{The authors contributed equally to this paper.}\orcidID{0000-0002-7421-3389} \and
Jan Portisch\inst{3,4}$^\star$\orcidID{0000-0001-5420-0663} \and
Heiko Paulheim\inst{4}\orcidID{0000-0003-4386-8195} \and
Harald Sack\inst{1,2}\orcidID{0000-0001-7069-9804} \and
Mehwish Alam\inst{1,2}\orcidID{0000-0002-7867-6612}}
\authorrunning{Biswas, Portisch, Paulheim, Sack, Alam}
\institute{
FIZ Karlsruhe -- Leibniz Institute for Infromation Infrastructure, Germany\\
\email{firstname.lastname@fiz-karlsruhe.de}\\
\and
Karlsruhe Institute of Technology, Institute AIFB, Germany
\and
SAP SE, Walldorf, Germany\\
\email{jan.portisch@sap.com}\\
\and
Data and Web Science Group, University of Mannheim, Germany\\
\email{\{jan,heiko\}@informatik.uni-mannheim.de}}
\maketitle             
\begin{abstract}
The entity type information in Knowledge Graphs (KGs) such as DBpedia, Freebase, etc. is often incomplete due to automated generation or human curation. Entity typing is the task of assigning or inferring the semantic type of an entity in a KG. This paper presents \textit{GRAND}, a novel approach for entity typing leveraging different graph walk strategies in RDF2vec together with textual entity descriptions. RDF2vec first generates graph walks and then uses a language model to obtain embeddings for each node in the graph. This study shows that the walk generation strategy and the embedding model have a significant effect on the performance of the entity typing task. The proposed approach outperforms the baseline approaches on the benchmark datasets DBpedia and FIGER for entity typing in KGs for both fine-grained and coarse-grained classes. The results show that the combination of order-aware RDF2vec variants together with the contextual embeddings of the textual entity descriptions achieve the best results.

\keywords{Entity Type Prediction  \and RDF2vec \and Knowledge Graph Embedding \and Graph Walks \and Language Models}
\end{abstract}
%
%
%

%%%%%%%%%%%%%%%%%%%%%%%%
%Main matter 
%%%%%%%%%%%%%%%%%%%%%%%%
\section{Introduction}
Many efforts have been made towards the automated generation of Knowledge Graphs (KGs) from heterogeneous resources such as text or images. One such effort is the creation of cross-domain KGs such as DBpedia~\cite{auer2007dbpedia}, Wikidata~\cite{vrandevcic2014wikidata}, Freebase~\cite{bollacker2008freebase}, etc. which are either extracted automatically from structured data, generated using heuristics, or are human-curated. This leads to incomplete information in the KGs which can occur on factual level (e.g., missing entities and/or relations between the entities) or on schema level (e.g., the missing entity type information). For instance, DBpedia version 2016-10 consists of 48 subclasses of \textit{dbo:Person}; however, only 36.6\% of the total number of entities belonging to \textit{dbo:Person} are assigned to its subclasses. Moreover, 307,164 entities in the entire DBpedia 2016-10 version are assigned to \textit{owl:Thing}. 

\begin{figure}[t!]
\caption{Excerpt from DBpedia}
\includegraphics[width=\textwidth]{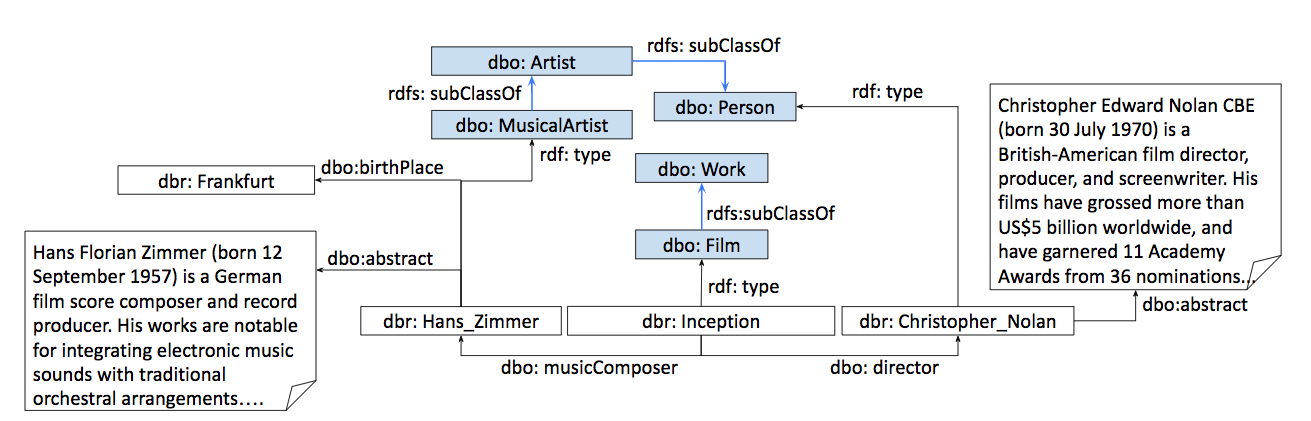}
  
  \label{fig:intro}
\end{figure}

To address the KG incompleteness on the factual level, a lot of models~\cite{bordes2011learning,dettmers2018conve,schlichtkrull2018modeling}, etc. have been proposed. These models focus mainly on predicting the missing entities and relations in the KGs but not the entity types. However, the entity type information in KGs plays a vital role in various Natural Language Processing based applications such as question answering~\cite{tong2019leveraging}, relation extraction~\cite{jain2018type}, recommendation, or system~\cite{wang2019explainable}. Following these lines, this paper focuses on the problem of entity typing which is the task of assigning or inferring the semantic type of an entity in a KG. Figure~\ref{fig:intro} shows an excerpt from DBpedia where the class \textit{dbo:MusicalArtist}%\footnote{prefix dbo: \url{http://dbpedia.org/ontology/}} 
is a subclass of \textit{dbo:Artist} which is a subclass of \textit{dbo:Person}. \textit{dbo:Artist} and \textit{dbo:MusicalArtist}, respectively, are the fine-grained entity types for \textit{dbr:Hans\_Zimmer}
%\footnote{prefix dbr: \url{http://dbpedia.org/resource/}} 
and \textit{dbo:Artist} is the missing type information. \textit{dbo:Person} is the coarse-grained type.

Recent years have witnessed a few studies on entity typing approaches in KGs using heuristics~\cite{paulheim2013type} and machine learning based classification models~\cite{melo2016type,YaghoobzadehAS18,jin-etal-2018-attributed,jin-etal-2019-fine,BiswasSSA21}. These models predict entity types using different KG features such as the anchor text mentions in the textual entity descriptions, relations between the entities, entity names, and Wikipedia categories. They learn the representation of the entities from their KG structure by using translational models~\cite{lin2015learning}, GCN-based models~\cite{jin-etal-2019-fine}, neighborhood based attention models~\cite{zhuo2022neighborhood} followed by the correlation between the entities and its types. These models exploit the neighborhood information only by the entities directly connected, i.e., the triple information of the entities. However, the large amount of contextual information of the entities captured in the graph walks remains unexplored. The work presented in this paper emphasizes on modeling the KG by taking advantage of the semantics of graph walks to predict the entity types with the help of different kinds of walk generation strategies, such as classic random walks, entity walks, and property walks. The paths generated by these graph walk strategies are used within the RDF2vec model~\cite{DBLP:journals/semweb/RistoskiRNLP19} to generate different entity representations. 
Additionally, the textual entity descriptions in the KGs contain rich semantic information which is beneficial in predicting the missing entity types. For instance, as depicted in Figure~\ref{fig:intro}, the textual entity descriptions of the entities clearly mentions that \textit{dbr: Christopher\_Nolan} is a \textit{director}, \textit{dbr: Hans\_Zimmer} is a \textit{music composer}, and \textit{dbr: Inception} is a \textit{film}. Some of the existing baseline models such as MuLR~\cite{SchutzeY17} use non-contextual Neural Language Models (NLMs), whereas the other uses GCN model~\cite{jin-etal-2019-fine} on the words extracted from the entity descriptions. Therefore, to capture the contextual information of the textual entity description contextual NLM, is used to generate entity representations.  

This paper presents a framework named \textbf{GRAND} (\textbf{G}raph Walks for \textbf{R}DF2vec \textbf{a}nd E\textbf{n}tity \textbf{D}escriptions), which exploits different variants of the RDF2vec model based on different graph walk strategies together with textual entity descriptions to predict the missing entity types in a KG. In this work, the entity typing problem is modelled as a classification problem. A flat and a hierarchical classification model are deployed on the top of the feature vectors generated from the aforementioned entity representations to predict the missing entity types. The empirical results based on the extensive experiments on two benchmark datasets FIGER~\cite{YaghoobzadehAS18} and DBpedia630k~\cite{ZhangZL15} show that the proposed approach is robust and outperforms the state-of-the-art (SOTA) models. Further experiments show that \textit{GRAND} performs considerably well on unseen entities. The main contributions of this work are:
\begin{itemize}
    \item A framework which leverages different graph walk strategies based RDF2vec models and a contextual NLM for textual entity descriptions is proposed to predict the missing entity types. 
    \item A generalized classification framework consisting of three different modules namely multi-class, multi-label, and hierarchical classification is introduced to predict the missing entity types on different levels of granularity. It can be easily deployed for predicting entity types on entity representations from any KGs.  
    \item Extensive experiments are conducted on the benchmark datasets to study the impact of several combinations of entity representations generated from the RDF2Vec variants and the NLM. An analysis on the weights in the classification has been conducted for analyzing which entity representations are suitable in which entity typing situations. Furthermore, the impact of dimensionality reduction of the entity representations on the local and global level using Principle Component Analysis (PCA) is studied.
\end{itemize}
The rest of the paper is organized as follows: Section~\ref{sec:relatedwork} gives an overview of the baseline approaches. Section~\ref{sec:approach} describes the proposed methodology, followed by experiments and results in Section~\ref{sec:experiments}. Finally, Section~\ref{sec:conclusion} provides the conclusion and an outlook of future work.

%Following these lines, this paper presents , an RDF2vec-based type prediction system for KGs. GRAND is capable of hierarchical type prediction by exploiting multiple classifiers that are ordered in a row. The inputs to the system are embedding representations of the concepts for which the types shall be predicted. GRAND supports multiple RDF2vec embedding variations. Several experiments have been conducted which study the effect of six different RDF2vec variants, i.e., three different walk generation strategies in combination with two embedding algorithms. The impact of vector fusion on the performance of ENTIRE has also been studied. The results are then compared to state-of-the-art type prediction systems where ENTIRE outperforms the baseline models significantly. Finally, an analysis on the the weights in the neural network has been conducted for analyzing which variants are suitable in which entity typing situations.

 %The work presented in this paper emphasizes on taking into account this path information of an entity with the help of different kinds of walk generation strategies.

%The non-contextual NLMs are static in nature and are context independent. The latent representation of the words do not dynamically change according to the context the words appear in. However, contextual embeddings encode semantics of the words differently based on different contexts.
\section{Related Work}
\label{sec:relatedwork}

This section discusses existing literature on entity typing and categorizes them based on their underlying methodology such as heuristics-based methods or machine learning based methods. 

SDType~\cite{paulheim2013type} is a statistical heuristic model that exploits links between instances using weighted voting. The model is based on the assumption that certain relations occur only with particular types. SDType often does not perform well if two or more classes share the same sets of properties and also if specific relations are missing for the entities.

One of the recent models, Cat2Type~\cite{BiswasSSA21}, takes into account the semantics underlying the textual information in the Wikipedia categories using language models such as BERT. In order to consider the structural information of Wikipedia categories, a category-category network is generated which is then fed to Node2Vec for obtaining the category embeddings. The embeddings of both structural and textual information are combined for classifying entities into their types. In~\cite{biswas2020entity}, different word embedding models, trained on triples, are leveraged together with a classification model to predict the entity types. Therefore, contextual information is not captured. In CUTE~\cite{XuZLXHW16}, a hierarchical classification model has been proposed which helps in cross-lingual entity typing by exploiting category, property, and property-value pairs. Another model has been proposed in~\cite{melo2016type} which performs type prediction using the Scalable Local Classifier per Node (SLCN) algorithm based on a set of incoming and outgoing relations. However, the entities with few relations are likely to be misclassified. MuLR~\cite{SchutzeY17} learns multi-level representations of entities via character, word, and entity embeddings followed by the hierarchical multi-label classification. Another model, namely FIGMENT~\cite{YaghoobzadehAS18}, uses a global model and a context model. The global model predicts entity types based on the entity mentions from the corpus and the entity names. The context model calculates a score for each context of an entity and assigns it to a type. Therefore, it requires a large annotated corpus which is a drawback of the model. In APE~\cite{jin-etal-2018-attributed}, a partially labeled attribute entity-entity network is constructed containing structural, attribute, and type information for entities followed by deep neural networks to learn the entity embeddings. MRGCN~\cite{wilcke2020end} is a multi-modal message-passing network that learns end-to-end from the structure of KGs as well as from multimodal node features. In HMGCN~\cite{jin-etal-2019-fine}, the authors propose a GCN-based model to predict the entity types considering the relations, textual entity descriptions, and the Wikipedia categories. ConnectE~\cite{zhao2020connecting} and AttET~\cite{zhuo2022neighborhood} models find correlation between neighborhood entities to predict the missing types. However, unlike GRAND, these two models do not look for information far away from the source entity. They work on the principal of L2 distance in their embedding space to detect the types and therefore not compared with the proposed model. Also, in order to employ the hidden layer as latent features for entity representation, restricted Boltzman machines (RBMs) are used to learn a target distribution across the usage of relations of entities~\cite{weller2021predicting}.

\section{Entity Type Prediction: GRAND framework}
\label{sec:approach}

\begin{figure}[t!]
    \centering
    \includegraphics[width=\textwidth]{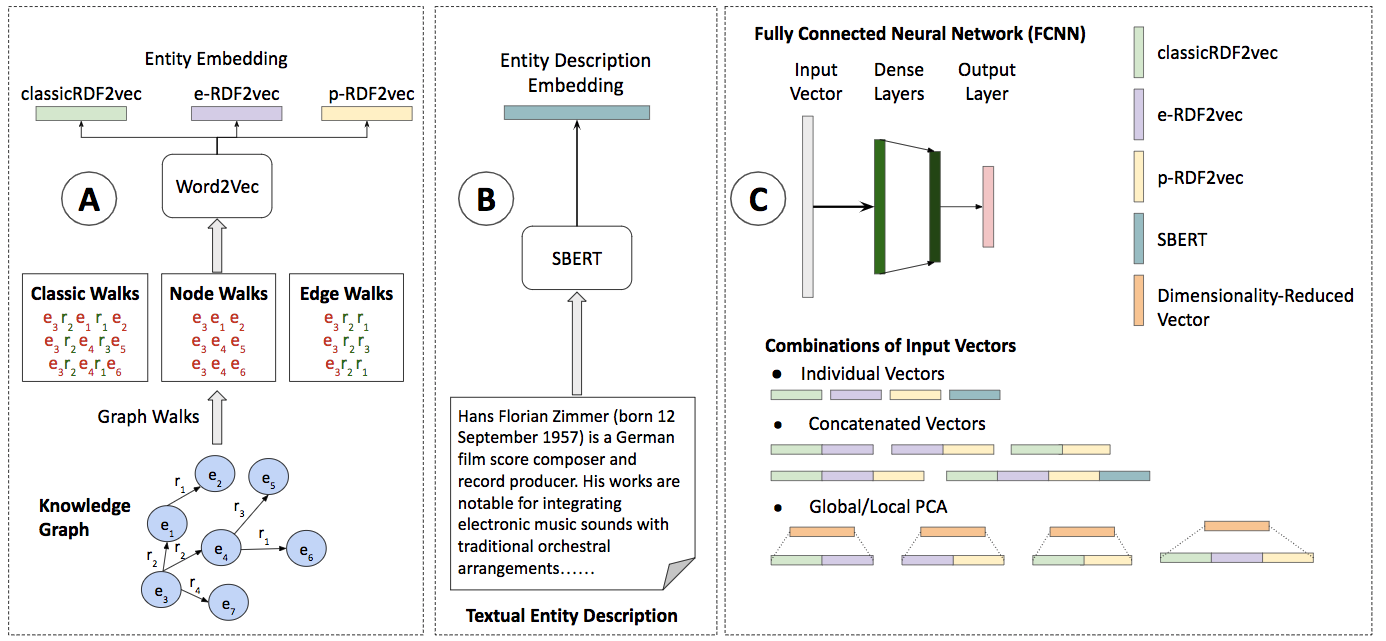}
    \caption{Architecture of the GRAND framework}
    \label{fig:grand_architecture}
\end{figure}

An overview of the GRAND framework is illustrated in Figure~\ref{fig:grand_architecture}. Component \encircle{A} represents the RDF2vec variants that use the different strategies for generating graph walks, i.e., classic walks, node walks, and property walks. %These walks are then given as an input to the language models separately and finally variants of entity representations are generated. 
Component \encircle{B} generates the representations of the entities from the textual entity description by using SBERT. Finally, component \encircle{C} shows combinations of the variants of entity representations used for flat as well as hierarchical classification. The rest of the section contains the explanation of the component details. 

\noindent\textbf{Preliminaries.} 
We define a knowledge graph $\mathcal{G}$ as a labeled directed graph $\mathcal{G} = (\mathcal{V},\mathcal{E})$, where $\mathcal{E} \subseteq \mathcal{V}\times\mathcal{R}\times\mathcal{V}$ for a set of relations $\mathcal{R}$. Vertices are subsequently also referred to as \emph{entities} and edges as \emph{predicates}.

%A KG $G$ consists a set of entities $E$ and its corresponding types are denoted by a set of types or classes $C_T$.  . \texttt{rdf:type} is used to define that an entity $e_i \in E$ is an instance of a class $c_k \in C_T$ and is given by a triple $<e_i,\texttt{rdf:type},c_k>$. \texttt{rdf:type} is an instance of \texttt{rdf:Property}.

%comes in two flavors: If there is merely one entity type to predict, ENTIRE will use one machine learning model which uses the RDF2vec concept vector representation as input. In the case of hierarchical type systems, i.e. if there are coarse and fine-grained types, ENTIRE will use one machine learning model per level. Each model uses the corresponding RDF2vec vector as input. This approach is known as \emph{local classifier per level} and is further described in Subsection~\ref{ssec:hierarchical_et_prediction}. 

%This section discusses RDF2vec variants with different walk generation strategies. It then explains the overall architecture of the proposed approach ``ENTIRE" along with the vector configurations and finally provides details on the classifier for entity type prediction. In the following, a KG $\mathcal{G}$ can be defined as a labeled directed graph with vertices and edges, i.e., $\mathcal{G} = (\mathcal{V},\mathcal{E})$, where $\mathcal{E} \subseteq \mathcal{V}\times \mathcal{V} \times \mathcal{R}$ for a set of relations $\mathcal{R}$.

\subsection{Entity Representation}
\label{ssec:rdf2vec_variations}
RDF2vec~\cite{DBLP:journals/semweb/RistoskiRNLP19} is one of the first approaches to adopt statistical language modeling techniques to KGs. The key idea of RDF2vec is a two-step approach: first, random walks over the graph are executed, thereby collecting sequences of entities and relations. To employ language modeling techniques, these sequences are then considered as sentences where each entity and relation in the sequence are treated as words. In RDF2vec, those sentences are then processed by word2vec~\cite{mikolov2013distributed,mikolov2013efficient}, where both variants of word2vec, i.e., continuous bag of words (CBOW) and skip-gram (SG), are possible.

One limitation of the word2vec algorithm is that it is not aware of the word order. For instance, for a window size of 4, the sentences ``John ate a pizza'' and ``pizza ate a John'' are equivalent. This is also the case with \emph{RDF2vec}: For instance, the statements \texttt{<Severus> <loves> <Lily>} and \texttt{<Lily> <loves> <Severus>}, are considered equivalent even though \texttt{<loves>} is not a symmetric property. To overcome this limitation, an order-aware version of RDF2vec has been proposed~\cite{DBLP:conf/semweb/PortischP21} which has shown improved performance on multiple machine learning datasets. This order-aware variant of RDF2vec uses a  \textit{structured word2vec} model~\cite{DBLP:conf/naacl/LingDBT15} which incorporates the positional information of the words in a sentence. The main advantage of the order-aware RDF2vec model over the classical RDF2vec model is that it respects the positional information of the entities and relations in the random walks, thereby learning embeddings which are better in terms of type separation.

Another type of RDF2vec extension is to explore different strategies for performing graph walks. These strategies have been explored using either variants of random walks (e.g., community hops~\cite{keikha2018community}, walklets~\cite{perozzi2017don}, or hierarchical walks~\cite{schlotterer2019investigating}), or by combining different random walk strategies, as the \emph{ontowalk2vec} approach, which combines RDF2vec and node2vec walks~\cite{gkotse2020ontology}.
In this paper, the aforementioned order-aware as well as different RDF2vec graph walk strategies~\cite{DBLP:journals/corr/abs-2204-02777} are leveraged to predict the missing types of the entities. 

\noindent\textbf{Graph Walk Generation Strategies.} RDF2vec combines the notion of similarity and relatedness. This can be easily observed when printing the most related concepts for ``Berlin'' on DBpedia via KGvec2go~\cite{DBLP:conf/lrec/PortischHP20}, i.e., many people who are \emph{related} to the city are identified as politicians. However, those are not really \emph{similar} -- they do not share properties with Berlin (which is a city rather than a living being). This leads to further exploration of RDF2vec for entity typing.

In this paper, six different RDF2vec configurations are presented and evaluated -- stand alone as well as combinations.
For the task of entity typing, three different walk generation strategies are applied: (1) classic walks, (2) entity walks, and (3) predicate walks. Each strategy is explained below in more detail.

\noindent\textbf{\textit{Classic Walks.}}
The originally presented RDF2vec variant generates multiple random walks for each node in the graph.
A random walk of length $n$ (where $n$ is an even number) is of the form
\begin{equation}
    w = (w_{-\frac{n}{2}}, w_{-\frac{n}{2}+1}, ..., w_0,..., w_{\frac{n}{2}-1}, w_\frac{n}{2})
\end{equation}
where $w_i \in \mathcal{V}$ if $i$ is even, and $w_i \in \mathcal{R}$ if $i$ is odd.
For better readability, we stylize $w_i \in \mathcal{V}$ as $e_i$ and $w_i \in \mathcal{R}$ as $p_i$:
\begin{equation}
    w = (e_{-\frac{n}{2}}, p_{-\frac{n}{2}+1}, ..., e_0,..., p_{\frac{n}{2}-1}, e_\frac{n}{2})
\end{equation}

\noindent\textbf{\textit{Entity Walks (e-RDF2vec).}} An entity walk contains only entities without any other properties. Such an approach is also known as \emph{e-RDF2vec}, given by
\begin{equation}
w_e = (e_{-\frac{n}{2}}, e_{-\frac{n}{2}+2}, ..., e_0,..., e_{\frac{n}{2}-2}, e_{\frac{n}{2}})
\end{equation}
For an entity walk, all elements are entities, i.e., $w_{n_i} \in \mathcal{V}$.

\noindent\textbf{\textit{Predicate Walks (p-RDF2vec).}}
A predicate walk contains only one entity together with object properties known as \emph{p-RDF2vec} and is defined as:
\begin{equation}
w_p = (p_{-\frac{n}{2}+1}, p_{-\frac{n}{2}+3}, ..., e_0,..., p_{\frac{n}{2}-3}, p_{\frac{n}{2}-1})
\end{equation}

\noindent The different walk strategies are visualized in component \encircle{\bf{A}} in Figure~\ref{fig:intro}.

\begin{comment}
\begin{figure}
    \centering
    \includegraphics[width=0.6\textwidth]{figures/walk_figure.png}
    \caption{Walk strategies exemplified for node \texttt{C} using a sample graph. Nodes are colored in blue, edges in red. }
    \label{fig:walk_strategies}
\end{figure}
\end{comment}

\noindent\textbf{Generating Entity Embeddings using RDF2Vec variants.}
An embedding model is trained for each set of walks using word2vec~\cite{mikolov2013distributed,mikolov2013efficient} and position-aware word2vec~\cite{DBLP:conf/naacl/LingDBT15} (suffix $oa$ in the following) which yields six sets of embeddings: (1) Classic RDF2vec, (2) e-RDF2vec, (3) p-RDF2vec, (4) Classic RDF2vec$_{oa}$, (5) e-RDF2vec$_{oa}$, and (6) p-RDF2vec$_{oa}$. 
The proposed model, GRAND, is evaluated by using the configurations presented in~\ref{ssec:rdf2vec_variations} on their own as well as in a fused way. Concerning the fusion of vectors, three modes are employed: (1) Vector concatenation, (2) Local PCA (LPCA), and (3) Global PCA (GPCA). PCA is a technique for reducing the dimensionality of the vectors with minimal loss in encoded information. It is used for identification of a smaller number of uncorrelated variables known as principal components. The difference between (2) and (3) is that in the case of the LPCA, a principal component analysis is only performed for the subset of vectors that appear in the datasets (see Section~\ref{ssec:datasets}) whereas for the GPCA, one all vectors generated from the KG using RDF2vec variants are considered. Each of these configurations can be used as vector within GRAND (see component \encircle{C} in Figure~\ref{fig:grand_architecture}).

The main advantages of using different RDF2vec variants are: \textbf{(i)} With a growing length of walks and training window, they can take advantage of large entity context ranges by effectively treating every entity as being connected to all the others in the graph -- this is in contrast to the baseline models which are based on local aggregation, i.e. they learn the representation of each entity based on its adjacent entities in the KG~\cite{jin-etal-2019-fine,zhuo2022neighborhood}. \textbf{(ii)} The graph walk strategies are effective, robust, and equitable, i.e., all relations and nodes are given equal importance in generating the embeddings. \textbf{(iii)} The walk strategies put emphasis on certain semantic aspects -- namely \emph{relatedness} and \emph{similarity}~\cite{DBLP:journals/corr/abs-2204-02777}. \textbf{(iv)} RDF2vec is a very scalable embedding algorithm. \textbf{(v)} The experimental results from~\cite{zouaq2020schema} show RDF2vec works better on the separability task compared to the other embedding models. The separability task aims at measuring if embeddings from different classes can be linearly separable and the evaluation is done on 10,000 pairs of classes from DBpedia.
\textbf{(vi)} Any classification algorithm can be deployed on top of  entity embeddings to predict the missing types. 

\subsection{Entity Description Representation}
The textual descriptions of an entity provide rich semantic information. Sentence-BERT (SBERT)~\cite{ReimersG19} fine-tunes the BERT~\cite{DevlinCLT19} model using the siamese and triplet networks to update the weights such that the resulting sentence embeddings are semantically meaningful and semantically similar sentences are closely positioned in the embedding space. For one epoch, a 3-way softmax classifier objective function is used for the fine-tuning of the BERT model. In the training phase of SBERT, two input sentences are passed through the BERT model followed by a pooling layer namely, MEAN-strategy, and MAX-strategy. A fixed-size representation for the input sentences are generated by this pooling layer. Next, they are concatenated with the element-wise difference and multiplied with a trainable weight. The cross-entropy loss is used for optimization. In order to encode the semantics, the twin network is fine-tuned on Semantic Textual Similarity data. SBERT model follows a two-step process in which it is first trained on Wikipedia via BERT and then fine-tuned on Natural Language Inference (NLI) data. NLI is a collection of 1,000,000 sentence pairs created by combining The Stanford Natural Language Inference (SNLI)
%\footnote{\url{https://nlp.stanford.edu/projects/snli/}} 
and Multi-Genre NLI (MG.NLI) datasets.

In this work, the same approach is followed to extract the embedding of the textual entity descriptions as mentioned in the evaluation of the quality of sentence embeddings in~\cite{ReimersG19}. Given be a textual entity description $D_{e_i}$ denoted by a sequence of words $\left\{W_1, W_2, ...,W_n\right\}$, where $W_j$ is the $j^{th}$ word in the entity description, and $e_i$ is the corresponding entity. The entity description $D_{e_i}$ is considered as a single sequence of words which is provided as an input to the SBERT model to get the embedding of the textual entity description ${\bf{E_{D_i}}}$. The pre-trained SBERT model used in GRAND is the SBERT-SNLI-STS-base model which is fine tuned on SNLI and STS datasets which outperforms the baseline models as shown in\cite{ReimersG19}. The MEAN pooling strategy is used in the pooling layer.

The main advantages of using pre-trained SBERT model are: \textbf{(i)} Since the pre-trained SBERT model is fine-tuned with two different datasets, the entity description embeddings obtained lose domain-specific knowledge and bias, and learn task-agnostic properties of the language. \textbf{(ii)} Unlike static word embedding models, such as word2vec, the contextual embedding model SBERT encodes semantics of the words differently based on different contexts. Therefore, the entity description embeddings capture the contextual information for the task of entity typing unlike the baseline models~\cite{jin-etal-2019-fine,SchutzeY17} \textbf{(iii)} They are computationally inexpensive as the model is pre-trained on huge amount of text and can be easily fine-tuned based on the information available. \textbf{(iv)} A representation of the entities can be obtained from the textual entity description for long-tailed entities in the KG, i.e., entities with no or few properties. \textbf{(v)} A task-specific classification model can be deployed on top of the entity description embeddings for entity typing task as illustrated in the proposed GRAND framework.

\subsection{Entity Type Prediction}
GRAND consists of three different classification modules: {\bf{(1)}} Multi-class, {\bf{(2)}} Multi-label, and {\bf{(3)}} Hierarchical, that are discussion below.

\noindent\textbf{Entity Representation.} The aforementioned approaches generate entity embeddings from various RDF2vec variants and from the contextual embedding model SBERT, which are provided as input to the classification modules. The input entity vectors are generated by concatenating the different vectors generated by the embedding models as depicted in component \encircle{C} in Figure~\ref{fig:grand_architecture}. 
\begin{comment}

Formally, they are given by:
\begin{equation}
\begin{split}
     E_{i} &= \bf{E_{RDF2vec_{classic}}^{V_i}} \oplus \bf{E_{p-RDF2vec}^{V_i}}  \oplus \bf{E_{e-RDF2vec}^{V_i}} \oplus \bf{E_{D_i}} \\
     &  =  \bf{E_{RDF2vec_{classic}}^{V_i}} \oplus \bf{E_{p-RDF2vec}^{V_i}}  \\
     &  =  \bf{E_{RDF2vec_{classic}}^{V_i}} \oplus \bf{E_{e-RDF2vec}^{V_i}}  \\
     &  =  \bf{E_{{p-RDF2vec}}^{V_i}} \oplus \bf{E_{e-RDF2vec}^{V_i}}  \\
     &  =  \bf{E_{p-RDF2vec}^{V_i}} \oplus \bf{E_{e-RDF2vec}^{V_i}}  \\
\end{split}
\end{equation}
\end{comment}

\noindent\textbf{Classifiers.} For \textit{\textbf{multi-class classification}}, a Fully Connected Neural Network (FCNN) consisting of two dense layers with ReLU as an activation function is deployed on the top of the entity representation. A softmax classifier with a cross-entropy loss function is used in the last layer to calculate the probability of the entities belonging to different classes. Formally it is given by,
\begin{equation}
    \label{eq:softmax}
     f(s)_{i} = \frac{e^{s_{i}}}{\sum_{j}^{C_{T}} e^{s_{j}}}, \quad\text{and}\quad  CE_{loss} = -\sum_{i}^{C_{T}}t_{i} log (f(s)_{i}),
\end{equation}
where $s_{j}$ are the scores inferred for each class in $C_{T}$ given in Equation~\ref{eq:softmax}. $t_i$ and $s_i$ are the ground truth and the score for each class in C, respectively.

\noindent In \textit{\textbf{multi-label classification}}, an entity can belong to more than one class or type. Therefore, a certain entity $e_i$ belonging to one class $c_i$ has no impact on the decision of it belonging to another class $c_j$, where $c_i, c_j \in C_T$.
A FCNN with RELU as an activation function is used for the two dense layers. A sigmoid function with binary cross-entropy loss is used in the last layer which sets up a binary classification problem for each class in $C_T$ and is given by,
\begin{equation}
      CE_{loss} = -t_{i} log(f(s_{i})) - (1 - t_{i}) log(1 - f(s_{i})),
\end{equation}
where $s_i$ and $t_i$ are the score and ground truth for $i^{th}$ class in $C_T$.

\noindent\textit{\textbf{Hierarchical Classification}} can be broadly categorized into local and global classification. The local information in local classifier can be utilized in different ways leading to different types of local classifiers such as Local classifier Per Node (LPN), a Local classifier Per Parent Node (LPPN) and a Local classifier Per Level (LPL)~\cite{SillaF11}. 
The proposed framework GRAND uses LPL which consists of training a flat classifier for each level of the class hierarchy. A multi-class classifier is trained at each level of the class hierarchy is used to discriminate among the classes at that level. The two main advantages of the LPL model are: \textbf{(i)} It is computationally efficient compared to LPN for large KGs consisting of large number of classes as LPN model would have equal number of classifiers. The number of classifiers in LPL are restricted to the number of levels in the class hierarchy. \textbf{(ii)} Since a single classifier is trained at each level, it reduces the horizontal class prediction inconsistencies. In GRAND, a two-layered FCNN with ReLU activation function and cross-entropy loss has been deployed at each level of the class hierarchy. However, one of the drawbacks of LPL is that an entity can be classified as class $1$ at one level and then it can be again classified as class $2.1$ on the second level. Here, class $2.1$ is not a subclass of $1$ and the entity should be classified to a subclass of $1$. In order to tackle such inconsistencies, in this work, the entity which is misclassified as $2.1$ in level $2$ will be typed as $1$ as its entity type as it was correctly identified in level $1$. 
\section{Experiments and Results}
\label{sec:experiments}
This section provides details on the benchmark datasets, experimental setup, analysis of the results obtained, and the ablation study.

%\subsection{Datasets}
\label{ssec:datasets}

\noindent\textbf{Datasets.} The two benchmark datasets FIGER~\cite{YaghoobzadehAS18} and DBpedia630k~\cite{ZhangZL15} are used to evaluate the performance of the GRAND framework against the baseline models. DBpedia630k consists of 630,000 entities and 14 non-overlapping classes and FIGER consists of 201,933 entities with 102 classes from Freebase. The entities of the extended DBpedia630k dataset are split equally into three parts DB-1, DB-2, and DB-3, each containing 210,000 entities. Each DBpedia split is divided into a train, test and validation set with 50\%, 30\%, and 20\% of the total entities respectively~\cite{jin-etal-2019-fine} as well as to 48 classes in the class hierarchy. There are no shared entities between the train, test, and validation sets for all the DBpedia630k splits and in FIGER. FIGER has been extended with triples from DBpedia as explained in~\cite{jin-etal-2019-fine,BiswasSSA21}. The statistics is provided in Table~\ref{tab:dataset-stat}. The code, and data are publicly available\footnote{\url{shorturl.at/abJRW}}. 

\begin{table}[t!]
\centering
\caption{Statistics of the datasets}
\scriptsize

%\resizebox{\textwidth}{!}{%
\begin{tabular}{l|c|c|c|c}\hline
Parameters                            & DB-1    & DB-2    & DB-3  & FIGER    \\ \hline
\#Entities                            & 210,000 & 210,000 & 210,000  & 201,933 \\
%\#Categories                          & 232,112 &  231,979 &  231,580  & 322,654 \\
\#Entities train                      & 105,000 & 105,000 & 105,000   & 101,266  \\
\#Entities test                       & 63,000  & 63,000  & 63,000 & 60,447  \\
\#Entities validation                 & 42,000  & 42,000  & 42,000 & 40,220     \\

\hline
\end{tabular}%
%}

\label{tab:dataset-stat}
\end{table}

\noindent\textbf{Experimental Setup.}
The experiments are conducted on six sets of embeddings: (1) Classic RDF2vec, (2) e-RDF2vec, (3) p-RDF2vec, (4) Classic RDF\-2vec$_{oa}$, (5) e-RDF2vec$_{oa}$, and (6) p-RDF2vec$_{oa}$. The walks are generated with a depth of 8 and 500 walks per entity. Classic and OA embeddings are trained using SG with 200 dimensions and 5 epochs. For training the order aware variants (4-6), walks from the corresponding non-order aware variants (1-3) are reused. The training was performed using the jRDF2vec framework\footnote{\url{https://github.com/dwslab/jRDF2Vec}}~\cite{DBLP:conf/semweb/PortischHP20}. %The RDF2vec embeddings are calculated on a Debian 11 machine with 768GiB of RAM and 32 cores \`a 2.60GHz (Intel Xeon). 
All the classifiers are used with the batch size 64, 100 epochs, and adam optmizer. %The SBERT model and classification models are performed on an Ubuntu 16.04.5 LTS system with 503GiB RAM with TITAN X (Pascal) GPU. 
The vectors are publicly available.\footnote{\url{https://bit.ly/3besaWF}}

\noindent\textbf{Results.}
In order to evaluate the proposed approach against the baseline models, Micro-averaged $F_1$ ($Mi$-$F_1$) and Macro-averaged $F_1$ ($Ma$-$F_1$) metrics are used along with the accuracy. Different variants of \textit{RDF2vec} have been evaluated which serve as an ablation study. The baselines used for the experiments are: CUTE~\cite{XuZLXHW16}, MuLR~\cite{SchutzeY17}, FIGMENT~\cite{YaghoobzadehAS18}, APE~\cite{jin-etal-2018-attributed}, HMGCN~\cite{jin-etal-2019-fine}, and CAT2Type~\cite{BiswasSSA21}. The results of the proposed framework on two benchmark datasets and their comparison with the baseline models are depicted in Table~\ref{tab:rdf2vec_sbert}.

\begin{table}[t!]
\centering
\scriptsize
\caption{Results of GRAND on benchmark datasets. The best result of each mode is printed in bold, the runner-up is underlined.}
\label{tab:rdf2vec_sbert}
\begin{tabular}{l|l|ll|ll|ll|ll} 
\hline
\multirow{2}{*}{}& \multirow{2}{*}{Model} & \multicolumn{2}{l|}{DB-1} & \multicolumn{2}{l|}{DB2}  & \multicolumn{2}{l|}{DB3}& \multicolumn{2}{l}{FIGER}\\ 
\cline{3-10}
 && Ma-F1 & Mi-F1  & Ma-F1  & Mi-F1  & Ma-F1 & Mi-F1 & Ma-F1 & Mi-F1           \\ 
\hline
\multirow{5}{*}{Baselines}& CUTE~\cite{XuZLXHW16} & 0.679& 0.702 & 0.681 & 0.713 & 0.685 & 0.717 & 0.743 & 0.782           \\
   & MuLR~\cite{SchutzeY17} & 0.748 & 0.771  & 0.757  & 0.784 & 0.752  & 0.775  & 0.776  & 0.812           \\
   & FIGMENT~\cite{YaghoobzadehAS18}  & 0.740 & 0.766  & 0.738  & 0.765   & 0.745 & 0.769  & 0.785 & 0.819           \\
   & APE~\cite{jin-etal-2018-attributed} & 0.758 & 0.784 & 0.761  & 0.785  & 0.760 & 0.782  & 0.722 & 0.756   \\
   & HMGCN-no hier~\cite{jin-etal-2019-fine}  & 0.785 & 0.812 & 0.794 & 0.820  & 0.791  & 0.817  & 0.789  & 0.827 \\
  % & HMGCN-hier  & 0.794  & 0.816  & 0.796 & 0.824  & 0.798 & 0.819  & 0.798  & 0.836           \\
   & CAT2Type-BERT~\cite{BiswasSSA21}  & \uline{0.983}  & \uline{0.984} & \uline{0.983} & \uline{0.983} & \uline{0.985}  & \uline{0.985}& \uline{0.764}  & \uline{0.881}           \\ 
\hline

\multirow{2}{*}{\begin{tabular}[c]{@{}l@{}}GRAND\\ Coarse-grained \end{tabular}}                         & \begin{tabular}[c]{@{}l@{}}classic-RDF2vec$_{oa}$ $\oplus$\\ s-RDF2vec$_{oa}$ $\oplus$\\ p-RDF2vec$_{oa}$ $\oplus$ SBERT  \end{tabular} & \bf{0.991}  & \bf{0.991}  & \bf{0.990} & \bf{0.990}  & \bf{0.989} & \bf{0.989}  & \bf{0.801} & \bf{0.893}           \\ 
\cline{2-10}
 & \begin{tabular}[c]{@{}l@{}}SBERT - only \end{tabular} & 0.972 & 0.972  & 0.97 & 0.97 & 0.97 & 0.97 & 0.648          & 0.844   \\

%\begin{tabular}[c]{@{}l@{}} GRAND\\  \end{tabular} & \begin{tabular}[c]{@{}l@{}}classic-RDF2vec$_{oa}$ $\oplus$\\ s-RDF2vec$_{oa}$ $\oplus$\\ p-RDF2vec$_{oa}$ $\oplus$ SBERT \end{tabular} & \bf{0.991}  & \bf{0.991}  & \bf{0.990} & \bf{0.990}  & \bf{0.989} & \bf{0.989}  & \bf{0.801} & \bf{0.893}           \\ 
\cline{1-2}\cline{3-10}

\multicolumn{1}{l|}{\multirow{2}{*}{\begin{tabular}[c]{@{}l@{}}Baselines\\Fine-grained\end{tabular}}} & \multicolumn{1}{l|}{CAT2Type-BERT~\cite{BiswasSSA21} } & 0.402 &  \multicolumn{1}{l|}{ 0.732} & 0.369   & \multicolumn{1}{l|}{0.721} & 0.847 & \multicolumn{1}{l|}{0.915} & 0.703 & 0.835           \\
\multicolumn{1}{l|}{}   & \multicolumn{1}{l|}{CAT2Type-node2vec~\cite{BiswasSSA21} }                                                                   & 0.391 & \multicolumn{1}{l|}{0.694} & 0.365 & \multicolumn{1}{l|}{0.677} & 0.807          & \multicolumn{1}{l|}{0.878} & 0.701          & 0.833           \\ 
\cline{1-2}\cline{3-10}
\begin{tabular}[c]{@{}l@{}}GRAND\\ Fine-grained \end{tabular}  & \begin{tabular}[c]{@{}l@{}}classic-RDF2vec$_{oa}$ $\oplus$\\ s-RDF2vec$_{oa}$ $\oplus$\\ p-RDF2vec$_{oa}$ $\oplus$ SBERT \end{tabular} & \bf{0.745} & \bf{0.870} & \bf{0.723} & \bf{0.851} & \bf{0.880}  & \bf{0.931}  & \bf{0.706}  & \bf{0.881}           \\ 
\hline
\begin{tabular}[c]{@{}l@{}}Baseline\\ Hierarchical \end{tabular}  & \begin{tabular}[c]{@{}l@{}}HMGCN-hier~\cite{jin-etal-2019-fine}  \end{tabular} & \bf{0.794}  & 0.816  & \bf{0.796} & 0.824  & \bf{0.798} & 0.819  & \bf{0.798}  & 0.836            \\ 
\hline

\multirow{2}{*}{\begin{tabular}[c]{@{}l@{}}GRAND\\ Hierarchical \end{tabular}}                         & \begin{tabular}[c]{@{}l@{}}classic-RDF2vec$_{oa}$ $\oplus$\\ s-RDF2vec$_{oa}$ $\oplus$\\ p-RDF2vec$_{oa}$ \end{tabular}  & \underline{0.731}  & \bf{0.882} & \underline{0.729} & \bf{0.881} & 0.726 & 0.877  & 0.701  & \underline{0.880}  \\ 
\cline{2-10}
 & \begin{tabular}[c]{@{}l@{}}classic-RDF2vec$_{oa}$ $\oplus$\\ s-RDF2vec$_{oa}$ $\oplus$\\ p-RDF2vec$_{oa}$ $\oplus$ SBERT \end{tabular} & \underline{0.731}                                     & 0.875                                                         & 0.718                                     & 0.869                     & \bf{0.935}          & \bf{0.946}                     & \underline{0.712}          & \bf{0.883}   \\
\hline
\end{tabular}
\end{table}

\noindent The results of GRAND as depicted in Table~\ref{tab:rdf2vec_sbert} can be obtained as follows: \textbf{(i)} \textit{Coarse-grained setting:} For DBpedia splits, the original dataset consisting of 14 non-overlapping classes is used. For FIGER, the number of coarse-grained classes is 30 and they are non-overlapping as well. Since, none of the entities belong to more than one class, \textit{multi-class} classification settings have been used here. \textbf{(ii)} \textit{Fine-grained setting:} The original DBpedia630k dataset is expanded with the DBpedia hierarchy to 37 fine-grained classes and these are non-overlapping classes. Therefore, a \textit{multi-class} classification model is used here as well. On the other hand, the FIGER dataset consists of overlapping fine-grained classes, i.e., one entity can belong to multiple classes. Therefore, a \textit{multi-label} classification is used for fine-grained FIGER dataset. \textbf{(iii)} For \textit{Hierarchical Classification}, a classifier on each level of the hierarchy is deployed. For DBpedia splits, it is a multi-class classification model and for FIGER it is a multi-label classification model at each level of the hierarchy. The baseline models which use a non-hierarchical classification such as CAT2Type~\cite{BiswasSSA21} also use a multi-class classification for DBpedia splits and a multi-label one for FIGER dataset. The SOTA model for hierarchical classification HMGCN~\cite{jin-etal-2019-fine} uses multi-label classification model.
The results show that GRAND outperforms the SOTA model CAT2Type with an improvement of 0.8\% on $Ma$-$F_1$ and 0.7\% on $Mi$-$F_1$ for DB-1, 0.7\% and 0.4\% on both the metrics for DB-2 and DB-3 respectively for the coarse-grained classes. The original dataset with 14 classes which do not contain the hierarchy is used for this coarse-grained non-hierarchical variant. Furthermore, for hierarchical classification, the proposed model significantly outperforms the SOTA HMGCN-hier model with an increment of 6.6\% for DB-1, 5.7\% for DB-2, and 12.7\% for DB-3 on the $Mi$-$F_1$ measure. For FIGER, the coarse-grained approach is a multi-class classification whereas the fine-grained approach is a multi-label classification. GRAND achieves the best results for FIGER on the coarse-grained approach which outperforms the baseline models. Moreover, with the multi-label fine-grained settings it achieves comparable results with the non-hierarchical baseline model CAT2Type and significantly outperforms the other non-hierarchical model HMGCN. One advantage of GRAND over CAT2Type is that it can be applied to any KGs and is not restricted to KGs containing information on Wikipedia Categories.
Table~\ref{tab:coarse-grained-results} and Table~\ref{tab:fine-grained-results} show the experimental results of the proposed approach for the coarse-grained and fine-grained classes respectively with different variants of RDF2vec and their combinations. %The experiments were conducted on coarse-grained classes and fine-grained classes referred to as \emph{Coarse-Grained} and \emph{Fine-Grained} respectively in the tables. 
%The values in bold represent the best results, the underlined values are the second best results, and the values which are in bold and underlined are the best results overall. 
%\emph{Single} represents the experimental results of the single trained model, \emph{Concat} represents the concatenation of the models. Finally, \emph{LPCA} and \emph{GPCA} refers to the results on Local PCA and Global PCA respectively. 
Experiments using \emph{Single} strategy show that \emph{all} order-aware RDF2vec embeddings significantly outperform their classic counterparts. Hence, the fusion strategies only focus on position-aware embeddings reducing the combinatorial complexity.

\noindent{\bf Impact of RDF2vec variants on Coarse-Grained Entity Typing.}
Table~\ref{tab:coarse-grained-results} shows the results of the experiment for coarse-grained entity typing. On the DB1 Split of the dataset, the best results for GRAND are obtained where the models are combined, i.e., $classic$-$RDF2vec_{oa}\oplus p$-$RDF2vec_{oa}\oplus e$-$RDF2vec_{oa}~(concat)$ outperforms $HMGCN$ for $Ma$-$F_1$ by 0.1744 and for $Mi$-$F_1$ by 0.148 and achieves comparable results with CAT2Type. However, e-RDF2vec configurations perform the weakest on their own but introduces additional value when combined with other approaches as depicted in the concat model. The best performing configuration includes the entity embeddings. Given the data, it appears that the PCA discards too much valuable information for DBpedia splits but not for FIGER. Overall, it can be observed that the performance differences between p-RDF2vec and classic-RDF2vec are minor. Nonetheless, the embeddings encode different information which is visible when combining the embeddings. Therefore, it can be concluded that the contextual information of the entities in form of path captures the characteristics features of the entities. Similar observation has been made for both DB2, DB3 split and FIGER. A detailed analysis of the impact of different vector components is provided in Section~\ref{sec:vector_analysis}.

\begin{sidewaystable}
\setlength{\tabcolsep}{1pt}
\caption{Evaluation of Single Classifier Results on the Coarse-Grained Dataset. The best result of each mode is printed in bold, the runner-up is underlined. The overall best configuration for each dataset is bold and underlined.}
\footnotesize
\begin{tabular}{|l|l|l|lll|lll|lll|lll|}

\hline
\multirow{2}{*}{\textbf{Dataset}} & \multirow{2}{*}{\textbf{Mode}} & \multirow{2}{*}{\textbf{Model}}                                                                                             & \multicolumn{3}{l|}{\textbf{DB-1}}                                                                  & \multicolumn{3}{l|}{\textbf{DB-2}}                                                & \multicolumn{3}{l|}{\textbf{DB-3}}
& \multicolumn{3}{l|}{\textbf{FIGER}}
\\ \cline{4-15} 
&   &  & \multicolumn{1}{l|}{ACC} & \multicolumn{1}{l|}{Ma-$F_1$} & Mi-$F_1$ & \multicolumn{1}{l|}{ACC} & \multicolumn{1}{l|}{Ma-$F_1$} & Mi-$F_1$ & 
                                                                                     \multicolumn{1}{l|}{ACC} & \multicolumn{1}{l|}{Ma-$F_1$} & Mi-$F_1$ &
                                                                                     \multicolumn{1}{l|}{ACC} & \multicolumn{1}{l|}{Ma-$F_1$} & Mi-$F_1$
                                                                                     \\ \hline
\multirow{23}{*}{\textbf{\begin{tabular}[c]{@{}l@{}}Coarse-\\ Grained\end{tabular}}} & \multirow{6}{*}{Single}        & classic-RDF2vec                                                                                                             & \multicolumn{1}{l|}{0.9163}            & \multicolumn{1}{l|}{0.9150}            & 0.9163            & \multicolumn{1}{l|}{0.9062}         & \multicolumn{1}{l|}{0.9043}                  & 0.9062         & \multicolumn{1}{l|}{0.9123}         & \multicolumn{1}{l|}{0.9109}                  & \multicolumn{1}{l|}{0.9123} 
& \multicolumn{1}{l|}{\underline{0.931}} & \multicolumn{1}{l|}{\bf{0.431}} & \multicolumn{1}{l|}{0.778} \\ \cline{3-15} 
                                                                                     &                                & classic-RDF2vec$_{oa}$                                                                                                      & \multicolumn{1}{l|}{{\bf 0.9448}}            & \multicolumn{1}{l|}{{\bf 0.9439}}            & {\bf 0.9448}            & \multicolumn{1}{l|}{{\bf 0.9346}}         & \multicolumn{1}{l|}{{\bf 0.9330}}                  & {\bf 0.9346}         & \multicolumn{1}{l|}{{\bf0.9457}}         & \multicolumn{1}{l|}{{\bf0.9449}}                  &  {\bf 0.9457} & \multicolumn{1}{l|}{\underline{\bf{0.933}}} & \multicolumn{1}{l|}{0.419} & \multicolumn{1}{l|}{\bf{\underline{0.781}}} \\ \cline{3-15} 
                                                                                     &                                & e-RDF2vec                                                                                                                & \multicolumn{1}{l|}{0.7352}            & \multicolumn{1}{l|}{0.7318}            & 0.7352            & \multicolumn{1}{l|}{0.7250}         & \multicolumn{1}{l|}{0.7308}                  & 0.7250         & \multicolumn{1}{l|}{0.7357}         & \multicolumn{1}{l|}{0.7304}                  &    0.7357      & \multicolumn{1}{l|}{0.927} & \multicolumn{1}{l|}{0.421} & \multicolumn{1}{l|}{0.771} \\ \cline{3-15} 
                                                                                     &                                & e-RDF2vec$_{oa}$                                                                                                         & \multicolumn{1}{l|}{0.7665}            & \multicolumn{1}{l|}{0.7651}            & 0.7665            & \multicolumn{1}{l|}{0.7625}         & \multicolumn{1}{l|}{0.7453}                  & 0.7625         & \multicolumn{1}{l|}{0.7694}         & \multicolumn{1}{l|}{0.7650}                  &    0.7694      & \multicolumn{1}{l|}{0.927} & \multicolumn{1}{l|}{0.422} & \multicolumn{1}{l|}{0.771} \\ \cline{3-15}  
                                                                                     &                                & p-RDF2vec                                                                                                                   & \multicolumn{1}{l|}{0.8949}            & \multicolumn{1}{l|}{0.8946}            & 0.8949            & \multicolumn{1}{l|}{0.8999}         & \multicolumn{1}{l|}{0.8914}                  & 0.8999         & \multicolumn{1}{l|}{0.8882}         & \multicolumn{1}{l|}{0.887}                  &  0.8882        & \multicolumn{1}{l|}{0.922} & \multicolumn{1}{l|}{\underline{0.426}} & \multicolumn{1}{l|}{0.778} \\ \cline{3-15} 
                                                                                     &                                & p-RDF2vec$_{oa}$                                                                                                            & \multicolumn{1}{l|}{\underline{0.9412}}            & \multicolumn{1}{l|}{\underline{0.9404}}            & \underline{0.9412}            & \multicolumn{1}{l|}{\underline{0.9332}}         & \multicolumn{1}{l|}{\underline{0.9303}}                  &  \underline{0.9332}        & \multicolumn{1}{l|}{\underline{0.9430}}         & \multicolumn{1}{l|}{\underline{0.9421}}                  &    \underline{0.9430}      & \multicolumn{1}{l|}{0.928} & \multicolumn{1}{l|}{0.422} & \multicolumn{1}{l|}{\underline{0.779}} \\ \cline{2-15}  
                                                                                     & \multirow{4}{*}{Concat}        & \begin{tabular}[c]{@{}l@{}}e-RDF2vec$_{oa}$ \\ $\oplus$ p-RDF2vec$_{oa}$\end{tabular}                                    & \multicolumn{1}{l|}{0.9518}            & \multicolumn{1}{l|}{0.9512}            & 0.9518            & \multicolumn{1}{l|}{0.9482}         & \multicolumn{1}{l|}{0.9412}                  & 0.9482         & \multicolumn{1}{l|}{0.9502}         & \multicolumn{1}{l|}{0.9495}                  & 0.9502         & \multicolumn{1}{l|}{0.912} & \multicolumn{1}{l|}{0.414} & \multicolumn{1}{l|}{0.77} \\ \cline{3-15} 
                                                                                     &                                & \begin{tabular}[c]{@{}l@{}}e-RDF2vec$_{oa}$ \\ $\oplus$ classic-RDF2vec$_{oa}$\end{tabular}                              & \multicolumn{1}{l|}{0.9450}            & \multicolumn{1}{l|}{0.9444}            & 0.9450            & \multicolumn{1}{l|}{0.9450}         & \multicolumn{1}{l|}{0.9144}                  & 0.9450         & \multicolumn{1}{l|}{0.9452}         & \multicolumn{1}{l|}{0.9482}                  & 0.9452         & \multicolumn{1}{l|}{0.908} & \multicolumn{1}{l|}{0.418} & \multicolumn{1}{l|}{\underline{0.772}} \\ \cline{3-15} 
                                                                                     &                                & \begin{tabular}[c]{@{}l@{}}classic-RDF2vec$_{oa}$ \\ $\oplus$ p-RDF2vec$_{oa}$\end{tabular}                                 & \multicolumn{1}{l|}{\underline{0.9564}}            & \multicolumn{1}{l|}{\underline{0.9555}}            & \underline{0.9563}            & \multicolumn{1}{l|}{\underline{0.9560}}         & \multicolumn{1}{l|}{{\bf \ul{0.9546}}}                  & \underline{0.9560}         & \multicolumn{1}{l|}{{\bf \ul{0.9582}}}         & \multicolumn{1}{l|}{\underline{0.9513}}                  &  {\bf \ul{0.9592}}        & \multicolumn{1}{l|}{\underline{0.92}} & \multicolumn{1}{l|}{\bf{0.429}} & \multicolumn{1}{l|}{\bf{0.774}} \\ \cline{3-15} 
                                                                                     &                                & \begin{tabular}[c]{@{}l@{}}classic-RDF2vec$_{oa}$ \\ $\oplus$ p-RDF2vec$_{oa}$ \\ $\oplus$ e-RDF2vec$_{oa}$\end{tabular} & \multicolumn{1}{l|}{{\ul \textbf{0.9600}}}            & \multicolumn{1}{l|}{{\bf \ul{0.9594}}}            & {\bf \ul{0.9600}}            & \multicolumn{1}{l|}{{\bf \ul{0.9667}}}         & \multicolumn{1}{l|}{\underline{0.9544}}   & {\bf \ul{0.9667}}         & \multicolumn{1}{l|}{\underline{0.9572}}         & \multicolumn{1}{l|}{{\bf \ul{0.9564}}}                & 0.9574       & \multicolumn{1}{l|}{\bf{0.924}} & \multicolumn{1}{l|}{\underline{0.424}} & \multicolumn{1}{l|}{\underline{0.772}} \\ \cline{2-15} 
                                                                                     & \multirow{4}{*}{Local PCA}     & \begin{tabular}[c]{@{}l@{}}e-RDF2vec$_{oa}$ \\ $\oplus$ p-RDF2vec$_{oa}$\end{tabular}                                    & \multicolumn{1}{l|}{0.8855}            & \multicolumn{1}{l|}{0.8845}            & 0.8855            & \multicolumn{1}{l|}{0.8757}         & \multicolumn{1}{l|}{0.8770}                  & 0.8757         & \multicolumn{1}{l|}{0.8918}         & \multicolumn{1}{l|}{0.8905}                  & 0.8918         & \multicolumn{1}{l|}{\underline{0.921}} & \multicolumn{1}{l|}{\underline{0.422}} & \multicolumn{1}{l|}{0.769} \\ \cline{3-15}  
                                                                                     &                                & \begin{tabular}[c]{@{}l@{}}e-RDF2vec$_{oa}$ \\ $\oplus$ classic-RDF2vec$_{oa}$\end{tabular}                              & \multicolumn{1}{l|}{0.9323}            & \multicolumn{1}{l|}{0.9314}            & 0.9324            & \multicolumn{1}{l|}{0.9314}         & \multicolumn{1}{l|}{0.9122}                  & 0.9314         & \multicolumn{1}{l|}{0.9015}         & \multicolumn{1}{l|}{0.9000}                  & 0.9015         & \multicolumn{1}{l|}{0.919} & \multicolumn{1}{l|}{0.419} & \multicolumn{1}{l|}{\underline{0.770}} \\ \cline{3-15}  
                                                                                     &                                & \begin{tabular}[c]{@{}l@{}}classic-RDF2vec$_{oa}$ \\ $\oplus$ p-RDF2vec$_{oa}$\end{tabular}                                 & \multicolumn{1}{l|}{{\bf 0.9471}}            & \multicolumn{1}{l|}{{\bf 0.9466}}            & {\bf 0.9472}            & \multicolumn{1}{l|}{\underline{0.9442}}         & \multicolumn{1}{l|}{{\bf 0.9300}}                  & \underline{0.9442}         & \multicolumn{1}{l|}{\underline{0.9378}}         & \multicolumn{1}{l|}{\underline{0.9217}}                  & \underline{0.9378}         & \multicolumn{1}{l|}{0.92} & \multicolumn{1}{l|}{\underline{0.421}} & \multicolumn{1}{l|}{0.724} \\ \cline{3-15}  
                                                                                     &                                & \begin{tabular}[c]{@{}l@{}}classic-RDF2vec$_{oa}$ \\ $\oplus$ p-RDF2vec$_{oa}$ \\ $\oplus$ e-RDF2vec$_{oa}$\end{tabular} & \multicolumn{1}{l|}{\underline{0.9405}}            & \multicolumn{1}{l|}{\underline{0.9395}}            & \underline{0.9405}            & \multicolumn{1}{l|}{{\bf 0.9551}}         & \multicolumn{1}{l|}{\underline{0.9195}}                  & {\bf 0.9551}         & \multicolumn{1}{l|}{{\bf 0.9413}}         & \multicolumn{1}{l|}{{\bf 0.9402}}                  &    {\bf 0.9413}      & \multicolumn{1}{l|}{\bf{0.925}} & \multicolumn{1}{l|}{\bf{0.428}} & \multicolumn{1}{l|}{\bf{0.778}} \\ \cline{2-15}  
                                                                                     & \multirow{4}{*}{Global PCA}    & \begin{tabular}[c]{@{}l@{}}e-RDF2vec$_{oa}$ \\ $\oplus$ p-RDF2vec$_{oa}$\end{tabular}                                    & \multicolumn{1}{l|}{0.9325}            & \multicolumn{1}{l|}{0.9316}            & 0.9325            & \multicolumn{1}{l|}{0.9412}         & \multicolumn{1}{l|}{\underline{0.9330}}                  & \underline{0.9412}         & \multicolumn{1}{l|}{0.9321}         & \multicolumn{1}{l|}{	0.9310}                  &    0.9321      & \multicolumn{1}{l|}{0.923} & \multicolumn{1}{l|}{\bf{0.428}} & \multicolumn{1}{l|}{\underline{0.778}} \\ \cline{3-15}  
                                                                                     &                                & \begin{tabular}[c]{@{}l@{}}e-RDF2vec$_{oa}$ \\ $\oplus$ classic-RDF2vec$_{oa}$\end{tabular}                              & \multicolumn{1}{l|}{0.9413}            & \multicolumn{1}{l|}{0.9405}            & 0.9414            & \multicolumn{1}{l|}{0.9322}         & \multicolumn{1}{l|}{0.9311}                  & 0.9322         & \multicolumn{1}{l|}{0.9416}         & \multicolumn{1}{l|}{0.9405}                  &    0.9416      & \multicolumn{1}{l|}{0.925} & \multicolumn{1}{l|}{\underline{0.428}} & \multicolumn{1}{l|}{0.776} \\ \cline{3-15} 
                                                                                     &                                & \begin{tabular}[c]{@{}l@{}}classic-RDF2vec$_{oa}$ \\ $\oplus$ p-RDF2vec$_{oa}$\end{tabular}                                 & \multicolumn{1}{l|}{{\bf 0.9499}}            & \multicolumn{1}{l|}{{\bf 0.9490}}            & {\bf 0.9499}            & \multicolumn{1}{l|}{\underline{0.9356}}         & \multicolumn{1}{l|}{0.9212}                  & 0.9356         & \multicolumn{1}{l|}{{\bf 0.9490}}         & \multicolumn{1}{l|}{{\bf 0.9482}}                  & {\bf 0.9490} & \multicolumn{1}{l|}{\underline{0.927}} & \multicolumn{1}{l|}{0.427} & \multicolumn{1}{l|}{0.767} \\ \cline{3-15} 
                                                                                     &                                & \begin{tabular}[c]{@{}l@{}}classic-RDF2vec$_{oa}$ \\ $\oplus$ p-RDF2vec$_{oa}$ \\ $\oplus$ e-RDF2vec$_{oa}$\end{tabular} & \multicolumn{1}{l|}{\underline{0.9476}}            & \multicolumn{1}{l|}{\underline{0.9468}}            & \underline{0.9476}            & \multicolumn{1}{l|}{{\bf 0.9568}}         & \multicolumn{1}{l|}{{\bf 0.9412}}                  & {\bf 0.9568}         & \multicolumn{1}{l|}{\underline{0.9489}}         & \multicolumn{1}{l|}{\underline{0.9481}}                  & \underline{0.9489}          & \multicolumn{1}{l|}{\bf{0.929}} & \multicolumn{1}{l|}{\bf{\underline{0.433}}} & \multicolumn{1}{l|}{\bf{0.779}} \\ 
                                                                                \hline
\end{tabular}
\label{tab:coarse-grained-results}
\end{sidewaystable}

\begin{sidewaystable}
\setlength{\tabcolsep}{1pt}
\caption{Evaluation of Single Classifier Results on the Fine-Grained Dataset. The best result of each mode is printed in bold, the runner-up is underlined. The overall best configuration for each dataset is bold and underlined.}
\footnotesize
\begin{tabular}{|l|l|l|lll|lll|lll|lll|}
\hline
\multirow{2}{*}{\textbf{Dataset}}       & \multirow{2}{*}{\textbf{Mode}} & \multirow{2}{*}{\textbf{Model}}                                                                                             & \multicolumn{3}{l|}{\textbf{DB-1}}                                                                              & \multicolumn{3}{l|}{\textbf{DB-2}}                                                                  & \multicolumn{3}{l|}{\textbf{DB-3}}                                                                  
& \multicolumn{3}{l|}{\textbf{FIGER}}
\\ \cline{4-15} 
                                        &                                &                                                                                                                             & \multicolumn{1}{l|}{ACC}     & \multicolumn{1}{l|}{Ma-$F_1$}     & Mi-$F_1$     & \multicolumn{1}{l|}{ACC} & \multicolumn{1}{l|}{Ma-$F_1$} & \textbf{Mi-$F_1$} & \multicolumn{1}{l|}{\textbf{ACC}} & \multicolumn{1}{l|}{Ma-$F_1$} & \textbf{$Mi-F_1$} & \multicolumn{1}{l|}{ACC} & \multicolumn{1}{l|}{Ma-$F_1$} & Mi-$F_1$ \\ \hline
\multirow{18}{*}{\textbf{\begin{tabular}[c]{@{}l@{}}Fine-\\ Grained\end{tabular}}} & \multirow{6}{*}{Single}        & classic-RDF2vec                                                                                                             & \multicolumn{1}{l|}{0.6716}                & \multicolumn{1}{l|}{0.374}                & 0.672                & \multicolumn{1}{l|}{0.6635}                  & \multicolumn{1}{l|}{\underline{0.363}}                  &           0.663        & \multicolumn{1}{l|}{0.8402}                  & \multicolumn{1}{l|}{\underline{0.736}}                  &        0.840           & \multicolumn{1}{l|}{\bf{\underline{0.991}}} & \multicolumn{1}{l|}{\underline{0.467}} & \multicolumn{1}{l|}{\underline{0.774}} \\ \cline{3-15} 
                                        &                                & classic-RDF2vec$_{oa}$                                                                                                      & \multicolumn{1}{l|}{{\ul 0.704}}          & \multicolumn{1}{l|}{\textbf{0.386}}       & {\ul 0.704}          & \multicolumn{1}{l|}{\underline{0.701}}                  & \multicolumn{1}{l|}{	0.356}                  &  \underline{0.701}                & \multicolumn{1}{l|}{\bf 0.871}                  & \multicolumn{1}{l|}{\bf 0.774}                  &  {\bf 0.871}                 & \multicolumn{1}{l|}{\underline{0.987}} & \multicolumn{1}{l|}{\bf{0.469}} & \multicolumn{1}{l|}{\bf{0.778}} \\ \cline{3-15} 
                                        &                                & e-RDF2vec                                                                                                                & \multicolumn{1}{l|}{0.564}                & \multicolumn{1}{l|}{0.297}                & 0.5643                & \multicolumn{1}{l|}{0.5231}                  & \multicolumn{1}{l|}{0.3164}                  &       0.5231            & \multicolumn{1}{l|}{0.6709}                  & \multicolumn{1}{l|}{0.5632}                  &  0.6709 & \multicolumn{1}{l|}{0.946} & \multicolumn{1}{l|}{0.445} & \multicolumn{1}{l|}{0.721} \\ \cline{3-15} 
                                        &                                & e-RDF2vec$_{oa}$                                                                                                         & \multicolumn{1}{l|}{0.5831}                & \multicolumn{1}{l|}{0.3064}                & 0.5831                & \multicolumn{1}{l|}{0.5542}                  & \multicolumn{1}{l|}{	0.3174	}                  &        0.5442           & \multicolumn{1}{l|}{0.6926}                  & \multicolumn{1}{l|}{0.5747}                  &    0.6926               & \multicolumn{1}{l|}{0.951} & \multicolumn{1}{l|}{0.452} & \multicolumn{1}{l|}{0.722} \\ \cline{3-15} 
                                        &                                & p-RDF2vec                                                                                                                   & \multicolumn{1}{l|}{0.6500}                & \multicolumn{1}{l|}{0.3549}                & 0.6499                & \multicolumn{1}{l|}{0.6504}                  & \multicolumn{1}{l|}{0.3449}                  &        0.6504           & \multicolumn{1}{l|}{0.7848}                  & \multicolumn{1}{l|}{0.6513}                  & 0.7848                  & \multicolumn{1}{l|}{0.949} & \multicolumn{1}{l|}{\underline{0.467}} & \multicolumn{1}{l|}{0.77} \\ \cline{3-15}
                                        &                                & p-RDF2vec$_{oa}$                                                                                                            & \multicolumn{1}{l|}{\textbf{0.706}}       & \multicolumn{1}{l|}{{\ul 0.384}}          & \textbf{0.706}       & \multicolumn{1}{l|}{{\bf 0.702}}                  & \multicolumn{1}{l|}{	{\bf 0.381}	}                  &   {\bf    0.7022 }            & \multicolumn{1}{l|}{\underline{0.847}}                  & \multicolumn{1}{l|}{0.732}                  &  \underline{0.8471}                 & \multicolumn{1}{l|}{\underline{0.951}} & \multicolumn{1}{l|}{0.459} & \multicolumn{1}{l|}{0.772} \\ \cline{2-15} 
                                        & \multirow{4}{*}{Concat}        & \begin{tabular}[c]{@{}l@{}}e-RDF2vec$_{oa}$ \\ $\oplus$ p-RDF2vec$_{oa}$\end{tabular}                                    & \multicolumn{1}{l|}{0.699}                & \multicolumn{1}{l|}{0.378}                & 0.6996                & \multicolumn{1}{l|}{0.698}                  & \multicolumn{1}{l|}{0.388}                  &        0.698           & \multicolumn{1}{l|}{0.877}                  &  \multicolumn{1}{l|}{\underline{0.784}}                  & 0.877                  & \multicolumn{1}{l|}{0.949} & \multicolumn{1}{l|}{0.454} & \multicolumn{1}{l|}{0.774} \\ \cline{3-15} 
                                        &                                & \begin{tabular}[c]{@{}l@{}}e-RDF2vec$_{oa}$ \\ $\oplus$ classic-RDF2vec$_{oa}$\end{tabular}                              & \multicolumn{1}{l|}{0.698}                & \multicolumn{1}{l|}{0.374}                & 0.6978                & \multicolumn{1}{l|}{0.701}                  & \multicolumn{1}{l|}{0.384}                  &      	0.7011             & \multicolumn{1}{l|}{0.881}                  & \multicolumn{1}{l|}{0.7811}                  & 0.881                  & \multicolumn{1}{l|}{0.96} & \multicolumn{1}{l|}{\underline{0.512}} & \multicolumn{1}{l|}{\underline{0.781}} \\ \cline{3-15}
                                        &                                & \begin{tabular}[c]{@{}l@{}}classic-RDF2vec$_{oa}$ \\ $\oplus$ p-RDF2vec$_{oa}$\end{tabular}                                 & \multicolumn{1}{l|}{{\ul \textbf{0.707}}} & \multicolumn{1}{l|}{{\ul 0.386}}          & {\ul \textbf{0.707}} & \multicolumn{1}{l|}{\underline{0.719}}                  & \multicolumn{1}{l|}{{\bf \ul{0.396}}}                  &       \underline{0.719}            & \multicolumn{1}{l|}{\underline{0.887}}                  & \multicolumn{1}{l|}{0.781}                  & \underline{0.881}                  & \multicolumn{1}{l|}{0.955} & \multicolumn{1}{l|}{\bf{\underline{0.519}}} & \multicolumn{1}{l|}{0.778} \\ \cline{3-15} 
                                        &                                & \begin{tabular}[c]{@{}l@{}}classic-RDF2vec$_{oa}$ \\ $\oplus$ p-RDF2vec$_{oa}$ \\ $\oplus$ e-RDF2vec$_{oa}$\end{tabular} & \multicolumn{1}{l|}{{\ul 0.703}}          & \multicolumn{1}{l|}{{\ul \textbf{0.393}}} & {\ul 0.720}          & \multicolumn{1}{l|}{{\bf \ul{0.7204}}}                  & \multicolumn{1}{l|}{\underline{0.3912}}                  &        {\bf \ul{0.720}}           & \multicolumn{1}{l|}{{\bf \ul{0.890}}}                  & \multicolumn{1}{l|}{ {\bf \ul{0.801}}}                  &    {{\bf \ul{0.8908}}}              & \multicolumn{1}{l|}{\bf{0.961}} & \multicolumn{1}{l|}{\bf{\underline{0.519}}} & \multicolumn{1}{l|}{\bf{0.783}} \\ \cline{2-15}  
                                        & \multirow{4}{*}{Local PCA}     & \begin{tabular}[c]{@{}l@{}}e-RDF2vec$_{oa}$ \\ $\oplus$ p-RDF2vec$_{oa}$\end{tabular}                                    & \multicolumn{1}{l|}{0.653}                & \multicolumn{1}{l|}{0.358}                & 0.6538                & \multicolumn{1}{l|}{0.648}                  & \multicolumn{1}{l|}{\underline{0.385}}                  &         	0.648          & \multicolumn{1}{l|}{0.806}                  & \multicolumn{1}{l|}{0.695}                  & 0.8060                  & \multicolumn{1}{l|}{0.948} & \multicolumn{1}{l|}{0.457} & \multicolumn{1}{l|}{0.778} \\ \cline{3-15} 
                                        &                                & \begin{tabular}[c]{@{}l@{}}e-RDF2vec$_{oa}$ \\ $\oplus$ classic-RDF2vec$_{oa}$\end{tabular}                              & \multicolumn{1}{l|}{0.6865}                & \multicolumn{1}{l|}{0.3683}                & 0.6865                & \multicolumn{1}{l|}{0.6952	}                  & \multicolumn{1}{l|}{0.3682	}                  &         0.6952          & \multicolumn{1}{l|}{\underline{0.8746}}             & \multicolumn{1}{l|}{\underline{0.7770}}                  & \underline{0.8746}                  & \multicolumn{1}{l|}{0.951} & \multicolumn{1}{l|}{0.501} & \multicolumn{1}{l|}{0.779} \\ \cline{3-15} 
                                        &                                & \begin{tabular}[c]{@{}l@{}}classic-RDF2vec$_{oa}$ \\ $\oplus$ p-RDF2vec$_{oa}$\end{tabular}                                 & \multicolumn{1}{l|}{\textbf{0.7006}}       & \multicolumn{1}{l|}{\textbf{0.3902}}       & \textbf{0.7006}       & \multicolumn{1}{l|}{\underline{0.7116}}                  & \multicolumn{1}{l|}{\bf 0.3907}                  &        \underline{0.7116}           & \multicolumn{1}{l|}{\bf 0.8774	}                  & \multicolumn{1}{l|}{\bf 0.7801	}                  &      {\bf 0.8774}             & \multicolumn{1}{l|}{0.950} & \multicolumn{1}{l|}{0.504} & \multicolumn{1}{l|}{0.771} \\ \cline{3-15} 
                                        &                                & \begin{tabular}[c]{@{}l@{}}classic-RDF2vec$_{oa}$ \\ $\oplus$ p-RDF2vec$_{oa}$\\
                                        $\oplus$ e-RDF2vec$_{oa}$\end{tabular}    & \multicolumn{1}{l|}{{\ul 0.6936}}          & \multicolumn{1}{l|}{{\ul 0.3839}}          & {\ul 0.6936}          & \multicolumn{1}{l|}{\bf 0.7122	}                  & \multicolumn{1}{l|}{0.3438}                  &       {\bf 0.7123}           & \multicolumn{1}{l|}{0.864}                  & \multicolumn{1}{l|}{0.764}                  &                 0.864  & \multicolumn{1}{l|}{\bf{0.958}} & \multicolumn{1}{l|}{\underline{0.514}} & \multicolumn{1}{l|}{\underline{0.781}} \\ \cline{2-15} 
                                        & \multirow{4}{*}{Global PCA}    & \begin{tabular}[c]{@{}l@{}}e-RDF2vec$_{oa}$ \\ $\oplus$ p-RDF2vec$_{oa}$\end{tabular}                                    & \multicolumn{1}{l|}{0.6845}                & \multicolumn{1}{l|}{0.3716}                & 0.6844                & \multicolumn{1}{l|}{0.66125	}                  & \multicolumn{1}{l|}{0.3189	}                  &         0.6612          & \multicolumn{1}{l|}{0.855}                  & \multicolumn{1}{l|}{0.7525	}                  & 0.8547                  & \multicolumn{1}{l|}{0.942} & \multicolumn{1}{l|}{0.449} & \multicolumn{1}{l|}{\underline{0.772}} \\ \cline{3-15} 
                                        &                                & \begin{tabular}[c]{@{}l@{}}e-RDF2vec$_{oa}$ \\ $\oplus$ classic-RDF2vec$_{oa}$\end{tabular}                              & \multicolumn{1}{l|}{0.6908}                & \multicolumn{1}{l|}{\textbf{0.3879}}       & 0.6908                & \multicolumn{1}{l|}{0.67143}                  & \multicolumn{1}{l|}{	0.3119	}                  &        0.67143           & \multicolumn{1}{l|}{0.8677}                  & \multicolumn{1}{l|}{0.7686}                  &           0.8677        & \multicolumn{1}{l|}{0.945} & \multicolumn{1}{l|}{0.449} & \multicolumn{1}{l|}{0.769} \\ \cline{3-15} 
                                        &                                & \begin{tabular}[c]{@{}l@{}}classic-RDF2vec$_{oa}$ \\ $\oplus$ p-RDF2vec$_{oa}$\end{tabular}                                 & \multicolumn{1}{l|}{{\ul 0.6981}}          & \multicolumn{1}{l|}{{\ul 0.3778}}          & {\ul 0.6981}          & \multicolumn{1}{l|}{\underline{0.6881}}                  & \multicolumn{1}{l|}{\underline{0.3241}}                  &     	\underline{0.6881}              & \multicolumn{1}{l|}{\bf 0.8754}                  & \multicolumn{1}{l|}{\underline{0.7771}}                  & 	{\bf 0.8754}        & \multicolumn{1}{l|}{\underline{0.956}} & \multicolumn{1}{l|}{\underline{0.457}} & \multicolumn{1}{l|}{0.771} \\ \cline{3-15} 
                                        &                                & \begin{tabular}[c]{@{}l@{}}classic-RDF2vec$_{oa}$ \\ $\oplus$ p-RDF2vec$_{oa}$ \\ $\oplus$ e-RDF2vec$_{oa}$\end{tabular} & \multicolumn{1}{l|}{\textbf{0.7005}}       & \multicolumn{1}{l|}{0.3768}                & \textbf{0.7004}       & \multicolumn{1}{l|}{{\bf 0.7014}}                  & \multicolumn{1}{l|}{\bf 0.3228}                  &      {\bf 0.7014}             & \multicolumn{1}{l|}{\underline{0.8709}}                  & \multicolumn{1}{l|}{{\bf 0.7780}}                  & \underline{0.8709}                 & \multicolumn{1}{l|}{\bf{0.961}} & \multicolumn{1}{l|}{\bf{0.498}} & \multicolumn{1}{l|}{\bf{\underline{0.784}}} \\ \hline
\end{tabular}
\label{tab:fine-grained-results}
\end{sidewaystable}

\noindent{\bf Impact of RDF2vec variants on Fine-Grained Entity Typing.}
GRAND is compared with the two best variants of CAT2Type namely BERT and node2vec as shown in Table~\ref{tab:rdf2vec_sbert} and results show that the proposed model significantly outperforms the CAT2Type model for all DBpedia splits and FIGER. In general, it is observed for uneven class distribution the evaluation metric $Ma$-$F_1$ achieves lower values compared to 
$Mi$-$F_1$. However, the $Ma$-$F_1$ results of GRAND for DB1 and DB2 splits are much better than that of CAT2Type. It strengthens the fact that the representation of entities obtained using strategic graph walks and contextual embedding of  entity descriptions contain more information about entities compared to the embeddings used in CAT2Type.

%Since the baseline models have not been evaluated on the fine-grained entity types, it is not possible to compare our results to the baseline models. For DB1 split, $classic$-$RDF2vec_{oa} \oplus p$-$RDF2vec_{oa}$ and $classic$-$RDF2vec_{oa} \oplus p$-$RDF2vec_{oa}\oplus e$-$RDF2vec_{oa}$ outperform all of its variants in terms of $Mi$-$F_1$ and $Ma$-$F_1$ respectively. The differences in the performance of these two models is marginal. It is observed that the results of DB3 are consistent with DB1 split. However, for DB3 split, the results are better compared to the other two splits for all the combination variants of GRAND. This is because, the DB3 split is more balanced in comparison with the other two splits. 

\noindent{\bf Impact of RDF2vec on Hierarchical classification.} Table~\ref{tab:results-hierarchical-class} shows the results of the hierarchical classification of the GRAND framework on different levels of the class hierarchy. The performance is computed for only $classic$-$RDF2vec_{oa} \oplus p$-$RDF2vec_{oa}\oplus e$-$RDF2vec_{oa}$ since it is the highest performing model based on experiments discussed in previous sections. The results show higher performances on level 1 since the number of classes is lesser i.e., 5, as compared to other levels. %Comparable results are achieved on level 2 and level 3 with marginal differences since there is no significant difference in the number of classes on both levels (i.e., 11, 12). 
GRAND outperforms the baseline model $HMGCN$-$withHier$ for $Mi$-$F_1$ metric as depicted in Table~\ref{tab:rdf2vec_sbert}.

\begin{table}[t]
\footnotesize
\caption{Results of the GRAND-LPL classification model at each level}
\centering
%\resizebox{\textwidth}{!}{%
\begin{tabular}{c|c|ll|ll|ll}
\hline
 Level&\#classes & \multicolumn{2}{c|}{DB1} & \multicolumn{2}{c|}{DB2} & \multicolumn{2}{c}{DB3} \\ \cline{3-8}
 &   & $Ma$-$F_1$ & $Mi$-$F_1$  & $Ma$-$F_1$ & $Mi$-$F_1$ & $Ma$-$F_1$ & $Mi$-$F_1$ \\ \hline 

%HMGCN-withHier  &     & - &0.794 & 0.816 & &0.796  & 0.824& - & 0.798 & 0.819 \\ \hline
%Overall &    &   0.8815  &  0.7311   &  0.8815  &  0.8805  &  0.7287 &  0.8805   &  0.8771   & 0.7263  & 0.8771   \\ \hline
1 &  5    & {\bf 0.961}   &   {\bf 0.962}     &  {\bf 0.960} &   {\bf 0.960}     & {\bf 0.959} & {\bf 0.959}  \\ 
2 &  11  &  0.744   &  0.925   &  0.747  &  0.929 &  0.744  & 0.924 \\ 
3 &  12 & 0.857    &   0.934 &  0.851 &  0.926 & 0.859   & 0.935  \\ 
4 & 17 &   0.361  & 0.705 & 0.358 &  0.702 & 0.359    & 0.674  \\ \hline
%level 5     &      -     &     -      &    -    & 0.7024  & 0.3577 &  0.7025    &    0.6745 &   0.3600 & 0.67456  \\ \hline
 
\end{tabular}
%}
\label{tab:results-hierarchical-class}
\end{table}

\noindent\textbf{Impact of Textual Entity Descriptions} To analyze the impact of entity descriptions, a multi-class classification was performed on the entity embeddings generated from the SBERT model. %For DBpedia splits containing 14 classes it results in $Ma$-$F_1$ and $Ma$-$F_1$ score of 0.972 on DB-1, 0.97 on both DB-2 and DB-3. For FIGER, a multi-label classification model is used which results in a  $Ma$-$F_1$ score of 0.648 and $Ma$-$F_1$ score of 0.844, 
As shown in Table~\ref{tab:rdf2vec_sbert}, GRAND with only SBERT performs better than all the baseline models except CAT2Type. Therefore, it can be concluded that contextual embeddings using SBERT provide the necessary relevant information as compared to the triple-based baseline models.

%\noindent{\bf Comparison with Heuristic Based Methods.}
%GRAND is compared with the statistical heuristic based entity typing approach SDType~\cite{paulheim2013type} in Table~\ref{tab:SDType}. For this, the publicly available results of the SDType method\footnote{\url{https://bit.ly/3eggWP0}} on DBpedia have been used. However, only a small fraction of the entities are common between the available results and DBpedia test datasets as shown in the second column of Table~\ref{tab:SDType}. Therefore, a comparison with the whole dataset is not possible. The accuracy provided in Table~\ref{tab:SDType} is calculated based on the number of common entities and compared with the best performing variation of GRAND. The result shows that GRAND outperforms SDType model. 

\noindent{\bf{Analysis of Vector Component Weight.}}\label{sec:vector_analysis} %As discussed above, 
In the experiments, it can be seen that the concatenation of embeddings achieves the best result. Therefore, it is further evaluated (1) which components are the most and the least important for the predictions and (2) whether there is a difference in the weights given the coarse-grained and the fine-grained prediction tasks. 

\noindent{{\bf Experimental Setup.}} In order to analyze the weights each vector component receives in the neural network, a FCNN with one layer was trained on the combination of all ordered aware RDF2vec (depicted in 1st 2 rows in coarse-grained and 1st 2 rows in fine-grained in Table~\ref{tab:network-weights}) and also with SBERT. It is noted that the overall goal of this setup is to analyze how much weight each of the four vector groups receive. Therefore, the sum of absolute weights in the network given to each vector is calculated for the first, and the tenth epoch. 

\noindent{{\bf Results.}}
The relative weights can be found in Table~\ref{tab:network-weights}. It is observed that the highest overall impact is independent of the dataset, achieved using the p-RDF2vec embeddings. This is followed by the classic RDF2vec embeddings. The least impact is achieved by the e-RDF2vec embeddings. Interestingly, a weight-shift occurs when switching from the coarse-grained entity typing to fine-grained entity typing, i.e., it is visible that the classic and the entity embeddings are more important for fine-grained predictions. The results suggest that p-RDF2vec is helpful for coarse-grained type prediction -- an intuitive finding given that p-RDF2vec encodes structural similarity. However, the more fine-grained the task gets, the more important are the \emph{actual} neighbor vertices.

\begin{table}[t]
\footnotesize
\centering
\caption{Relative network weights of each vector component group for DB-1 split.}
\begin{tabular}{l|l|l|l|l|l}
\hline
\textbf{Dataset}         & \textbf{Epoch} &   \textbf{SBERT}   &   \textbf{Classic RDF2vec$_{oa}$} & \textbf{p-RDF2vec$_{oa}$} & \textbf{e-RDF2vec$_{oa}$} \\ \hline
\multirow{4}{*}{Coarse-Grained} 
& 1 & - & 35.5\%    & 44.4\%  & 20.0\%  \\ 
& 10 & - & 32.9\%    & 49.9\%  & 17.1\%  \\   \cline{2-6}
& 1 & 58.04\% & 14.6\%  & 16.28\% & 11.08\% \\
& 10     & 47.9\%  & 18.5\%          & 22.8\%    & 10.8\%    \\ \hline
%& 100 & 29.6\%    & 49.4\%  & 21.0\% \\ \hline
\multirow{4}{*}{Fine-Grained}   
& 1 & -  & 35.4\%    & 42.1\%  & 22.5\%  \\ 
& 10 & - & 33.6\%    & 46.4\%  & 20.0\% \\  \cline{2-6}
 & 1 & 56.7\%  & 15.36\% & 16.84\% & 11.1\%  \\
 & 10     & 51.19\% & 16.83\%         & 19.5\%    & 12.48\% \\  \hline
%& 100 & 29.8\%    & 48.7\%  & 21.6\% \\ \hline
\end{tabular}

\label{tab:network-weights}
\end{table}

%\subsection{Classifying Unseen Entities}

%\subsection{Ablation Study}
\section{Summary \& Future Directions}
\label{sec:conclusion}
This paper proposes a novel entity type prediction framework, named \textbf{GRAND} based on RDF2vec variants and textual entity descriptions. The variants are constructed by different walk generation strategies and a new order-aware variant of word2vec. GRAND is evaluated on DBpedia630k and FIGER datasets.  The results show that GRAND considerably outperforms all the baseline models. 
Also, given the weight analysis, further experimentation on more fine-granular type systems -- such as in YAGO~\cite{DBLP:conf/www/SuchanekKW07} or CaLiGraph~\cite{DBLP:conf/esws/HeistP20} is to be conducted.

%%%%%%%%%%%%%%%%%%%%%%%%
%References 
%%%%%%%%%%%%%%%%%%%%%%%%

\bibliographystyle{splncs04}
\bibliography{references}

\begin{thebibliography}{10}
\providecommand{\url}[1]{\texttt{#1}}
\providecommand{\urlprefix}{URL }
\providecommand{\doi}[1]{https://doi.org/#1}

\bibitem{auer2007dbpedia}
Auer, S., Bizer, C., Kobilarov, G., Lehmann, J., Cyganiak, R., Ives, Z.G.:
  Dbpedia: {A} nucleus for a web of open data. In: Aberer, K., Choi, K., Noy,
  N.F., Allemang, D., Lee, K., Nixon, L.J.B., Golbeck, J., Mika, P., Maynard,
  D., Mizoguchi, R., Schreiber, G., Cudr{\'{e}}{-}Mauroux, P. (eds.) The
  Semantic Web, 6th International Semantic Web Conference, 2nd Asian Semantic
  Web Conference, {ISWC} 2007 + {ASWC} 2007, Busan, Korea, November 11-15,
  2007, Lecture Notes in Computer Science, vol.~4825, pp. 722--735. Springer
  (2007). \doi{10.1007/978-3-540-76298-0\_52},
  \url{https://doi.org/10.1007/978-3-540-76298-0\_52}

\bibitem{biswas2020entity}
Biswas, R., Sofronova, R., Alam, M., Sack, H.: Entity type prediction in
  knowledge graphs using embeddings. arXiv pp. arXiv--2004 (2020)

\bibitem{BiswasSSA21}
Biswas, R., Sofronova, R., Sack, H., Alam, M.: Cat2type: Wikipedia category
  embeddings for entity typing in knowledge graphs. In: Gentile, A.L.,
  Gon{\c{c}}alves, R. (eds.) {K-CAP} '21: Knowledge Capture Conference, Virtual
  Event, USA, December 2-3, 2021. pp. 81--88. {ACM} (2021).
  \doi{10.1145/3460210.3493575}, \url{https://doi.org/10.1145/3460210.3493575}

\bibitem{bollacker2008freebase}
Bollacker, K., Evans, C., Paritosh, P., Sturge, T., Taylor, J.: Freebase: a
  collaboratively created graph database for structuring human knowledge. In:
  ACM SIGMOD international conference on Management of data (2008)

\bibitem{bordes2011learning}
Bordes, A., Weston, J., Collobert, R., Bengio, Y.: Learning structured
  embeddings of knowledge bases. In: Proceedings of the Twenty-Fifth AAAI
  Conference on Artificial Intelligence (2011)

\bibitem{dettmers2018conve}
Dettmers, T., Pasquale, M., Pontus, S., Riedel, S.: Convolutional 2d knowledge
  graph embeddings. In: Proceedings of the 32th AAAI Conference on Artificial
  Intelligence (2018)

\bibitem{DevlinCLT19}
Devlin, J., Chang, M., Lee, K., Toutanova, K.: {BERT:} pre-training of deep
  bidirectional transformers for language understanding. In: Conference of the
  North American Chapter of the Association for Computational Linguistics:
  Human Language Technologies (2019)

\bibitem{gkotse2020ontology}
Gkotse, B.: Ontology-based Generation of Personalised Data Management Systems:
  an Application to Experimental Particle Physics. Ph.D. thesis, Universit{\'e}
  Paris sciences et lettres (2020)

\bibitem{DBLP:conf/esws/HeistP20}
Heist, N., Paulheim, H.: Entity extraction from wikipedia list pages. In:
  Harth, A., Kirrane, S., Ngomo, A.N., Paulheim, H., Rula, A., Gentile, A.L.,
  Haase, P., Cochez, M. (eds.) The Semantic Web - 17th International
  Conference, {ESWC} 2020, Heraklion, Crete, Greece, May 31-June 4, 2020,
  Proceedings. Lecture Notes in Computer Science, vol. 12123, pp. 327--342.
  Springer (2020). \doi{10.1007/978-3-030-49461-2\_19},
  \url{https://doi.org/10.1007/978-3-030-49461-2\_19}

\bibitem{jain2018type}
Jain, P., Kumar, P., Chakrabarti, S., et~al.: Type-sensitive knowledge base
  inference without explicit type supervision. In: Proceedings of the 56th
  Annual Meeting of the Association for Computational Linguistics (Volume 2:
  Short Papers). pp. 75--80 (2018)

\bibitem{jin-etal-2018-attributed}
Jin, H., Hou, L., Li, J., Dong, T.: Attributed and predictive entity embedding
  for fine-grained entity typing in knowledge bases. In: 27th International
  Conference on Computational Linguistics (2018)

\bibitem{jin-etal-2019-fine}
Jin, H., Hou, L., Li, J., Dong, T.: Fine-grained entity typing via hierarchical
  multi graph convolutional networks. In: Empirical Methods in Natural Language
  Processing and the 9th International Joint Conference on Natural Language
  Processing (2019)

\bibitem{SillaF11}
Jr., C.N.S., Freitas, A.A.: A survey of hierarchical classification across
  different application domains. Data Min. Knowl. Discov.  \textbf{22}(1-2),
  31--72 (2011). \doi{10.1007/s10618-010-0175-9},
  \url{https://doi.org/10.1007/s10618-010-0175-9}

\bibitem{keikha2018community}
Keikha, M.M., Rahgozar, M., Asadpour, M.: Community aware random walk for
  network embedding. Knowledge-Based Systems  \textbf{148},  47--54 (2018)

\bibitem{lin2015learning}
Lin, Y., Liu, Z., Sun, M., Liu, Y., Zhu, X.: Learning entity and relation
  embeddings for knowledge graph completion. In: Twenty-ninth AAAI conference
  on artificial intelligence (2015)

\bibitem{DBLP:conf/naacl/LingDBT15}
Ling, W., Dyer, C., Black, A.W., Trancoso, I.: Two/too simple adaptations of
  word2vec for syntax problems. In: {NAACL} {HLT} 2015. pp. 1299--1304. ACL
  (2015)

\bibitem{melo2016type}
Melo, A., Paulheim, H., V{\"o}lker, J.: Type {P}rediction in {RDF} {K}nowledge
  {B}ases {U}sing {H}ierarchical {M}ultilabel {C}lassification. In: WIMS (2016)

\bibitem{mikolov2013efficient}
Mikolov, T., Chen, K., Corrado, G., Dean, J.: Efficient estimation of word
  representations in vector space. arXiv preprint arXiv:1301.3781  (2013)

\bibitem{mikolov2013distributed}
Mikolov, T., Sutskever, I., Chen, K., Corrado, G., Dean, J.: Distributed
  representations of words and phrases and their compositionality. arXiv
  preprint arXiv:1310.4546  (2013)

\bibitem{paulheim2013type}
Paulheim, H., Bizer, C.: Type {I}nference on {N}oisy {RDF} {D}ata. In: ISWC
  (2013)

\bibitem{perozzi2017don}
Perozzi, B., Kulkarni, V., Chen, H., Skiena, S.: Don't walk, skip! online
  learning of multi-scale network embeddings. In: Proceedings of the 2017
  IEEE/ACM International Conference on Advances in Social Networks Analysis and
  Mining 2017. pp. 258--265 (2017)

\bibitem{DBLP:conf/lrec/PortischHP20}
Portisch, J., Hladik, M., Paulheim, H.: Kgvec2go - knowledge graph embeddings
  as a service. In: Calzolari, N., B{\'{e}}chet, F., Blache, P., Choukri, K.,
  Cieri, C., Declerck, T., Goggi, S., Isahara, H., Maegaard, B., Mariani, J.,
  Mazo, H., Moreno, A., Odijk, J., Piperidis, S. (eds.) Proceedings of The 12th
  Language Resources and Evaluation Conference, {LREC} 2020, Marseille, France,
  May 11-16, 2020. pp. 5641--5647. European Language Resources Association
  (2020), \url{https://aclanthology.org/2020.lrec-1.692/}

\bibitem{DBLP:conf/semweb/PortischHP20}
Portisch, J., Hladik, M., Paulheim, H.: Rdf2vec light - {A} lightweight
  approachfor knowledge graph embeddings. In: Taylor, K.L., Gon{\c{c}}alves,
  R.S., L{\'{e}}cu{\'{e}}, F., Yan, J. (eds.) Proceedings of the {ISWC} 2020
  Demos and Industry Tracks: From Novel Ideas to Industrial Practice co-located
  with 19th International Semantic Web Conference {(ISWC} 2020), Globally
  online, November 1-6, 2020 {(UTC)}. {CEUR} Workshop Proceedings, vol.~2721,
  pp. 79--84. CEUR-WS.org (2020),
  \url{http://ceur-ws.org/Vol-2721/paper520.pdf}

\bibitem{DBLP:conf/semweb/PortischP21}
Portisch, J., Paulheim, H.: Putting rdf2vec in order. In: Seneviratne, O.,
  Pesquita, C., Sequeda, J., Etcheverry, L. (eds.) Proceedings of the {ISWC}
  2021 Posters, Demos and Industry Tracks: From Novel Ideas to Industrial
  Practice co-located with 20th International Semantic Web Conference {(ISWC}
  2021), Virtual Conference, October 24-28, 2021. {CEUR} Workshop Proceedings,
  vol.~2980. CEUR-WS.org (2021), \url{http://ceur-ws.org/Vol-2980/paper352.pdf}

\bibitem{DBLP:journals/corr/abs-2204-02777}
Portisch, J., Paulheim, H.: Walk this way! entity walks and property walks for
  rdf2vec. CoRR  \textbf{abs/2204.02777} (2022)

\bibitem{ReimersG19}
Reimers, N., Gurevych, I.: Sentence-bert: Sentence embeddings using siamese
  bert-networks. In: Proceedings of the 2019 Conference on Empirical Methods in
  Natural Language Processing and the 9th International Joint Conference on
  Natural Language Processing, {EMNLP-IJCNLP} (2019)

\bibitem{DBLP:journals/semweb/RistoskiRNLP19}
Ristoski, P., Rosati, J., Noia, T.D., Leone, R.D., Paulheim, H.: Rdf2vec: {RDF}
  graph embeddings and their applications. Semantic Web  \textbf{10}(4),
  721--752 (2019). \doi{10.3233/SW-180317},
  \url{https://doi.org/10.3233/SW-180317}

\bibitem{schlichtkrull2018modeling}
Schlichtkrull, M., Kipf, T.N., Bloem, P., Van Den~Berg, R., Titov, I., Welling,
  M.: Modeling relational data with graph convolutional networks. In:
  Proceedings of the European semantic web conference (2018)

\bibitem{schlotterer2019investigating}
Schl{\"o}tterer, J., Wehking, M., Rizi, F.S., Granitzer, M.: Investigating
  extensions to random walk based graph embedding. In: 2019 IEEE International
  Conference on Cognitive Computing (ICCC). pp. 81--89. IEEE (2019)

\bibitem{DBLP:conf/www/SuchanekKW07}
Suchanek, F.M., Kasneci, G., Weikum, G.: Yago: a core of semantic knowledge.
  In: Williamson, C.L., Zurko, M.E., Patel{-}Schneider, P.F., Shenoy, P.J.
  (eds.) Proceedings of the 16th International Conference on World Wide Web,
  {WWW} 2007, Banff, Alberta, Canada, May 8-12, 2007. pp. 697--706. {ACM}
  (2007). \doi{10.1145/1242572.1242667},
  \url{https://doi.org/10.1145/1242572.1242667}

\bibitem{tong2019leveraging}
Tong, P., Zhang, Q., Yao, J.: Leveraging domain context for question answering
  over knowledge graph. Data Science and Engineering  (2019)

\bibitem{vrandevcic2014wikidata}
Vrande{\v{c}}i{\'c}, D., Kr{\"o}tzsch, M.: Wikidata: a free collaborative
  knowledgebase. Communications of the ACM  (2014)

\bibitem{wang2019explainable}
Wang, X., Wang, D., Xu, C., He, X., Cao, Y., Chua, T.S.: Explainable reasoning
  over knowledge graphs for recommendation. In: Proceedings of the AAAI
  conference on artificial intelligence (2019)

\bibitem{weller2021predicting}
Weller, T., Acosta, M.: Predicting instance type assertions in knowledge graphs
  using stochastic neural networks. In: Proceedings of the 30th ACM
  International Conference on Information \& Knowledge Management. pp.
  2111--2118 (2021)

\bibitem{wilcke2020end}
Wilcke, W., Bloem, P., de~Boer, V., van’t Veer, R., van Harmelen, F.:
  End-to-end entity classification on multimodal knowledge graphs. arXiv
  (2020)

\bibitem{XuZLXHW16}
Xu, B., Zhang, Y., Liang, J., Xiao, Y., Hwang, S., Wang, W.: Cross-lingual type
  inference. In: Database Systems for Advanced Applications - 21st
  International Conference, {DASFAA} (2016)

\bibitem{YaghoobzadehAS18}
Yaghoobzadeh, Y., Adel, H., Sch{\"{u}}tze, H.: Corpus-level fine-grained entity
  typing. J. Artif. Intell. Res.  (2018)

\bibitem{SchutzeY17}
Yaghoobzadeh, Y., Sch{\"{u}}tze, H.: Multi-level representations for
  fine-grained typing of knowledge base entities. In: 15th Conference of the
  European Chapter of the Association for Computational Linguistics (2017)

\bibitem{ZhangZL15}
Zhang, X., Zhao, J.J., LeCun, Y.: Character-level convolutional networks for
  text classification. In: Advances in Neural Information Processing Systems
  28: Annual Conference on Neural Information Processing Systems (2015)

\bibitem{zhao2020connecting}
Zhao, Y., Zhang, A., Xie, R., Liu, K., Wang, X.: Connecting embeddings for
  knowledge graph entity typing. In: Proceedings of the 58th Annual Meeting of
  the Association for Computational Linguistics. pp. 6419--6428 (2020)

\bibitem{zhuo2022neighborhood}
Zhuo, J., Zhu, Q., Yue, Y., Zhao, Y., Han, W.: A neighborhood-attention
  fine-grained entity typing for knowledge graph completion. In: Proceedings of
  the Fifteenth ACM International Conference on Web Search and Data Mining. pp.
  1525--1533 (2022)

\bibitem{zouaq2020schema}
Zouaq, A., Martel, F.: What is the schema of your knowledge graph? leveraging
  knowledge graph embeddings and clustering for expressive taxonomy learning.
  In: Proceedings of the international workshop on semantic big data. pp.~1--6
  (2020)

\end{thebibliography}

\end{document}